\documentclass[10pt,twocolumn,letterpaper]{article}

 \usepackage[pagenumbers]{cvpr} 
\usepackage{makecell}
\newcolumntype{C}{>{\centering\arraybackslash}X}  
\usepackage{graphicx}
\usepackage{amsmath}
\usepackage{amssymb}
\usepackage{booktabs}
\usepackage{tabularx}
\usepackage{adjustbox}
\usepackage{subcaption}
\usepackage{array}
\usepackage{tikz}
\usepackage{nicematrix}
\usepackage{caption}
\usepackage{capt-of}

\usepackage{xcolor}
\definecolor{Class_0!}{RGB}{0, 0, 0}
\definecolor{Class_1!}{RGB}{255, 0, 0}
\definecolor{Class_2!}{RGB}{70, 130, 255}
\definecolor{Class_3!}{RGB}{255, 255, 0}
\definecolor{Class_4!}{RGB}{255, 0, 255}
\definecolor{Class_5!}{RGB}{0, 255, 255}

\usepackage{booktabs}         
\usepackage{multirow}         
\usepackage{colortbl}         
\usepackage{xcolor}           

\definecolor{cvprblue}{rgb}{0.21,0.49,0.74}
\usepackage[pagebackref,breaklinks,colorlinks,allcolors=cvprblue]{hyperref}

\def\paperID{0} 
\def\confName{0}

\title{ Electrolyzers-HSI: Close-Range Multi-Scene  Hyperspectral Imaging Benchmark Dataset}

\author{\textsuperscript{1,2}Elias~Arbash, \textsuperscript{1}Ahmed~Jamal~Afifi, \textsuperscript{3,1}Ymane~Belahsen, \textsuperscript{1}Margret~Fuchs, \textsuperscript{1}Pedram~Ghamisi\\
\textsuperscript{2}Paul~Scheunders, \textsuperscript{1}Richard~Gloaguen\\
\textsuperscript{1} Helmholtz-Zentrum Dresden-Rossendorf (HZDR) - \\ Helmholtz Institute Freiberg for Resource Technology (HIF), Freiberg, Germany\\
\textsuperscript{2}University of Antwerp, Antwerpen, Belgium \\
\textsuperscript{3}National School of Applied Sciences of Oujda, Oujda, Morocco\\
}

\begin{document}

\twocolumn[{
\maketitle
    \begin{center}
        \includegraphics[width=\textwidth]{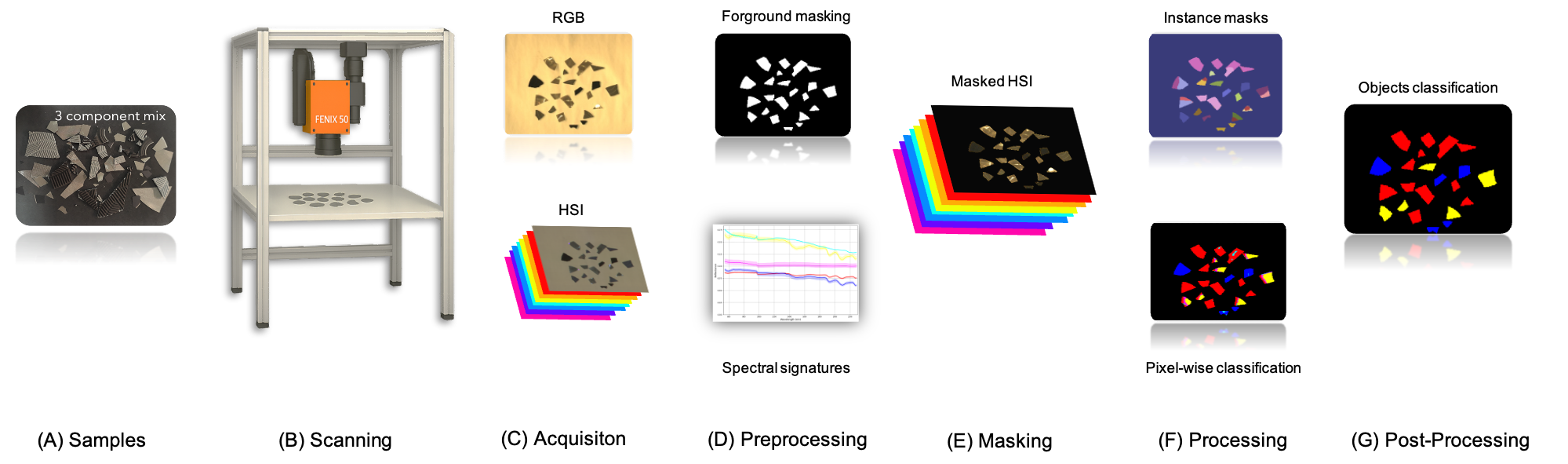}
        \scriptsize
    \colorbox{Class_1!}{Mesh}
    \colorbox{Class_2!}{Steel-Black}
    \colorbox{Class_3!}{Steel-Grey}
    \colorbox{Class_4!}{HTEL-Anode}
    \colorbox{Class_5!}{HTEL-Cathode}
        \captionof{figure}{An overview of the dataset acquisition and processing pipeline of the Electrolyzers-HSI dataset.
\textbf{(A)} The fragments of the shredded electrolyzer cells were organized into controlled scenes containing one to five material classes, enabling both isolated and mixed-material classification studies.
 \textbf{(B)} Each sample configuration was scanned on both sides using a dual-modality setup composed of high-resolution RGB and HSI sensors (400–2500nm), capturing refined spatial and rich spectral features.
 \textbf{(C)} The RGB images and HSI data cubes are acquired and coregistered.
 \textbf{(D)} The parallel preprocessing pipelines of the two modalities: reflectance conversion and normalization of HSI data cubes and zero-shot segmentation of the foreground objects in the RGB images.
 \textbf{(E) }HSI data background masking and foreground preservation.
 \textbf{(F)} Pixel-wise classification on the masked HSI and instance masks projection for majority voting.  
\textbf{(G)} Object-wise electrolyzers classification.} 
        \label{acquisition}
    \end{center}
}]

\begin{abstract}

The global challenge of sustainable recycling demands automated, fast, and accurate, state-of-the-art (SOTA)  material detection systems that act as a bedrock for a circular economy. Democratizing access to these cutting-edge solutions that enable real-time waste analysis is essential for scaling up recycling efforts and fostering the Green Deal. In response, we introduce \textbf{Electrolyzers-HSI}, a novel multimodal benchmark dataset designed to accelerate the recovery of critical raw materials through accurate electrolyzer materials classification. The dataset comprises 55 co-registered high-resolution RGB images and hyperspectral imaging (HSI) data cubes spanning the 400–2500 nm spectral range, yielding over 4.2 million pixel vectors and 424,169 labeled ones. This enables non-invasive spectral analysis of shredded electrolyzer samples, supporting quantitative and qualitative material classification and spectral properties investigation.  
We evaluate a suite of baseline machine learning (ML) methods alongside SOTA transformer-based deep learning (DL) architectures, including Vision Transformer, SpectralFormer, and the Multimodal Fusion Transformer, to investigate architectural bottlenecks for further efficiency optimisation when deploying transformers in material identification. We implement zero-shot detection techniques and majority voting across pixel-level predictions to establish object-level classification robustness. In adherence to the FAIR data principles, the electrolyzers-HSI dataset and accompanying codebase are openly available at \url{https://github.com/hifexplo/Electrolyzers-HSI} and \url{https://rodare.hzdr.de/record/3668}, supporting reproducible research and facilitating the broader adoption of smart and sustainable e-waste recycling solutions.

\end{abstract}    
\section{Introduction}
\label{sec:intro}

Hydrogen technology, particularly electrolyzers, receives focused attention due to its role in the energy transition strategy as a solution for energy transport and storage. Research and development put strong efforts into increasing the efficiency and upscaling of the main Electrolyzer types, which provides the perfect momentum to develop recycling strategies in parallel. The recovery of the valuable and critical resources contained in electrolyzers will contribute to securing electrolyzer raw material cycles, which support the sustainability of hydrogen-related strategies \cite{fleischhauer2015strength}. 

Electrolyzers recycling can benefit from HSI sensor technology along with SOTA ML and DL data processing models to precisely identify and recover critical materials, enhancing resource efficiency and circularity.
HSI sensors acquire detailed spectral information across hundreds of spectral bands, each reflecting unique interactions between the incident light and the material properties \cite{chang2003hyperspectral}. The non-invasive capability of HSI with the detailed spectral information from the scanned surface enables precise identification of different materials based on their spectral signatures. Apart from remote sensing applications, e.g., earth observation \cite{bioucas2013hyperspectral}, \cite{DLHSI}. Moreover, HSI offers essential capabilities for close-range sensing applications such as agriculture \cite{mishra2017close}, food \cite{foodHSI}, healthcare \cite{medicalHSI} and industry \cite{industrialHSI}.

Transformer-based DL models \cite{NIPS2017_3f5ee243}  have become a cornerstone of SOTA data processing methodologies, excelling in real-time performance and high accuracy criteria across diverse domains, including natural language processing with large language models such as ChatGPT \cite{gpt3, achiam2023gpt}, and computer vision, including both images \cite{CVT} and video \cite{videoT}. Recycling applications demand accurate, rapid, and dynamic solutions that greatly benefit from the application of HSI with these SOTA processing modalities that have end-to-end feature extraction capabilities when training data is abundant. This allows the detection of spectral features and patterns that can reveal unique material characteristics, supporting the decision-making in recycling facilities.

RGB images have high spatial resolution, emphasizing fine surface details, and their precise spatial features (e.g., traditional morphological features) provide clarified appearance-based characteristics for automated sorting in recycling streams. However, they may lack reliability for material identification due to appearance variations in the samples' end-of-life conditions. In contrast, material spectral features, derived from high spectral resolution HSI, offer a more robust criterion for accurate materials identification and classification. In this context, in-line, non-invasive scanning routines that combine high spatial resolution RGB with high spectral resolution HSI data emerge as a powerful solution. This multimodal approach not only enables the extraction of rich appearance features, which can further support precise point-wise validation, but also captures detailed spectral characteristics essential for material-wise investigation. When integrated within ground-breaking processing frameworks, these complementary modalities significantly boost detection performance, surpassing what can be achieved with either modality alone, and lay a strong foundation for reliable, scalable industrial recycling systems. This aligns with sustainable development and the circular economy, which seeks to enhance resource recovery, reduce waste generation, and supports global sustainability goal 12: Responsible Consumption and Production \cite{hak2016sustainable}, through recycling \cite{kirchherr2017conceptualizing, united2023transforming}.

\begin{table*}[ht]
    \centering
    \caption{Overview of HSI E-waste datasets.}
    \begin{adjustbox}{width=\textwidth}
    \begin{tabular}{l c c c c c }\toprule
        \textbf{Dataset}& \textbf{Size}& \textbf{Modality}& \textbf{Range}& \textbf{Sensor}&\textbf{Task(s)}  \\\midrule
        & & & & & \\
        Leone et al. \cite{ewaste}& +108 Point measurement& Hyperspectral vectors& 350 - 2500 nm& FieldSpec 4 spectroradiometer&Polymers classification\\
 & & & & & \\
                
        Tecnalia WEEE \cite{ewaste1} & 13 scenes& HSI & 400 – 1000 nm& Specim PHF Fast10 camera& E-waste metals segmentation\\
 & & & & & \\
        
        WEEE Plastic \cite{ewaste2}& Multiple plastic fragments scenes& HSI& 1000 – 2500 nm& Specim ImSpector N25E  &Polymers classification\\
 & & & & & \\
        
        Thermal E-Waste \cite{ewaste3}& Multiple E-waste IR scenes& HSI& 8000 – 15000 nm& FLIR ORION SC7000 & E-waste samples classification\\
 & & & & & \\
 Lambers et al. \cite{ewaste7}& 37 samples& HSI& 400 - 1000 nm& Innospec GreenEye& Color prediction of regranulate\\
 & & & & &\\
 Polymers \cite{arbash2024investigating} & 9 scenes & HSI& 380 - 2500 nm& Specim FENIX & Polymers classification\\
 & & & & &\\
 SpectralWaste \cite{ewaste6}& 852 labelled scene + 6803 unlabelled scenes& RGB + HSI& 1000 - 1700 nm& Teledyne DALSA Linea + Specim FX17&Ewaste samples segmentation\\
 & & & & &\\
 PCB-Vision \cite{arbash2024pcb}& 53 scenes& RGB + HSI& 400–1000 nm& Teledyne Dalsa C4020 + Specim FX10&PCB components segmentation\\
 & & & & & \\

        Electrolyzers-HSI& 55 scenes& RGB + HSI& 380 - 2500 nm& Teledyne Dalsa C4020 +  Specim FENIX& Electrolyzers classification\\
        
 & & & & &\\
De Lima Ribeiro et al. \cite{de2024multi}&  23 samples + 1 scene& Raman + HSI& 400–3400 $\text{cm}^{-1}$, 480-5300 nm& HORIBA ARAMIS Raman spectrometer, FENIX + FX50 &Polymers identification\\ \bottomrule
    \end{tabular}
    \end{adjustbox}
    \label{tab:E-waste datasets}
\end{table*}

In this study, we contribute to sustainability by optimising the decision-making routines in E-waste recycling streams for electrolyzers materials. We selected high-temperature electrolyzers (HTEL) because their components contain a range of high-tech and critical raw materials, i.e., rare-earth elements, Ni, Zr, and Mn contained in the Anode and Cathode ceramics of the HTEL, as well as relevant metals (frame, interconnectors, meshes).
Accordingly, HTEL represents a valuable source of secondary raw materials. Our core objective is to accurately detect major components in electrolyzers' recycling streams using non-invasive sensors combined with real-time data processing to support the decision-making of downstream E-waste recycling. For this reason, we provide a new multi-scene, multi-modality, high-resolution benchmark dataset of electrolyzers materials and investigate the performance of the native Transformer-based HSI processing models for the identification of electrolyzers materials. 
Additionally, we identify performance bottlenecks and highlight further optimisation directions.
Multi-scene HSI benchmark datasets are crucial for the development of DL models to ensure their generalizability across diverse scenes, since such datasets mimic industrial applications with continuous data acquisition (new HSI scenes) of mixed sample streams, over moving conveyor belts. High-quality and standardized datasets help models to learn robust and transferable representations, reducing the risk of overfitting. By exposing a model to numerous scenes and samples, it captures universal patterns and features inherent to the materials regardless of the different scenes. This directly improves its adaptability and reliable performance  across multiple domains and applications.
Our contribution in this work can be summarized as follows:
\begin{itemize}
    \item Introducing Electrolyzers-HSI: a dataset comprising 55 high spectral resolution HSI data cubes acquired in visible-near infrared (VNIR) and shortwave infrared (SWIR) ranges, each paired with their high spatial resolution co-registered RGB twin image and classification ground truth masks.
    \item Validating the dataset usage for electrolyzers identification with multiple standard ML and DL single- and multimodal Transformer-based models. 
    \item Enhancing the classification performance through object-level approaches, using zero-shot segmentation for background masking and foreground processing.
    \item Identifying performance limitations of Transformer-based architectures when applied to HSI data analysis. 
    \item Providing complete processing and inference pipeline codes and model weights for replication and deployment.
\end{itemize}

Fig. \ref{acquisition} presents an overview of the Electrolyzers-HSI workflow from sample preparation to the final object classification.
The remainder of the paper is organized as follows: Section 2 reviews related work, including the SOTA processing modalities and the available HSI E-waste datasets. Section 3 describes the dataset, its sample composition, along the statistical information. Section 4 presents the processing pipeline, methodologies, pixel- and object-level evaluations, with a discussion about the performance limitations and future directions. Section 5 concludes with a summary of key findings and contributions.

\section{Related Work}

With hundreds of spectral bands per pixel, HSI suffers from the curse of dimensionality and high redundancy. This complexity necessitates advanced, non-linear operations to effectively process the data, as traditional linear methods struggle to capture the intricate patterns. DL models, with their non-linear architectures, are well-suited for this task as they can uncover complex relationships within the data. As a consequence, they excel in extracting joint spatial-spectral features, making them ideal for HSI data processing \cite{DLHSI}.
Accordingly, several DL modalities were applied for HSI processing and classification, including fully connected network \cite{HSIFCN}, recurrent neural networks \cite{HSIRNN}, convolutional neural network (CNN) \cite{HSICNN, HSI3DCNN}, pure Transformers \cite{hong2021spectralformer, HSIT2}.  Hybrid implementations of Transformers with other techniques like CNN were utilized \cite{HSIhybrid, HSIhybrid2} to leverage the convolution mechanism for local spatial feature extraction together with the self-attention mechanism of Transformers, which enables the capture of both short- and long-range dependencies when plentiful data is provided.
. Following the trend in developing HSI processing models, we focus on pure Transformer-based models, especially the original implementation of Transformers in HSI \cite{hong2021spectralformer} avoiding further advanced variants like \cite{HSI-enhancedT, HSIT} in order to highlight the original architecture bottlenecks and performance limitations.

Data availability and quality are crucial for developing effective DL processing methods. However, HSI datasets for E-waste remain limited to only a few studies. Table \ref{tab:E-waste datasets} provides an overview of the referenced datasets, including the number of HSI scenes or data units reported by the authors, the available modalities, sensor names, spectral ranges, and the specific research objectives. Picon et al. provided Tecnalia WEEE, a 13-scene HSI dataset for the detection of different metallic E-waste samples, scanned in the VNIR \cite{ewaste1}. Bonifazi et al. \cite{ewaste2} explored SWIR HSI for polymer characterization to enhance plastic identification, aiding quality control and sorting processes in E-waste recycling streams. Using point measurement devices, Leone et al. \cite{ewaste} acquired a plastic hyperspectral reflectance dataset in the VNIR and SWIR spectral ranges from various samples, including pristine and degraded specimens. Gathering and utilising up to 9 HSI data cubes from the HSI scenes in \cite{de2024multi} \cite{de_lima_ribeiro_andrea_2023_2430}, in \cite{arbash2024investigating} , Arbash et al. investigate the performance of SOTA HSI classification models in polymers classification. Thermal E-Waste \cite{ewaste3} by Paulraj et al. is a thermal HSI dataset in the longwave infrared (LWIR) range for the classification of mixed E-waste samples (metals, plastics, PCBs, glass). Aiming at predicting the color of regranulates based on the color content of input flakes, Lambers et al. \cite{ewaste7} generated a hyperspectral dataset comprised of 185 measurements of 37 samples: 29 flakes and 8 colored samples. 
Several works demonstrate the value of multimodal RGB-HSI data for enhanced material characterization and sorting. PCB-Vision by Arbash et al. \cite{arbash2024pcb} is a dataset of 55 HSI-RGB scenes of printed circuit boards (PCBs) with segmentation ground truth for the PCB board and several main PCB components, including integrated circuits, Capacitors, and Connectors.
Casao et al. contributed with Spectral Waste \cite{ewaste6}, an RGB-HSI dataset containing 852 non-overlapping labeled and 6803 unlabeled scenes from an operational plastic waste sorting facility. The authors proposed a processing pipeline, integrating different DL modalities for general waste object segmentation, along with a co-registration method for the different modalities.
Extending further to Raman sensor, De Lima Riberio et al. \cite{de2024multi} characterize the improvements of polymer identification using Raman point measurement sensor and HSI.

\section{Dataset Description}
\label{sec:dataset_description}

\begin{figure}[!ht]
\centering
\begin{subfigure}[t]{0.45\columnwidth}
    \includegraphics[width=1\textwidth, height = 4.2cm]{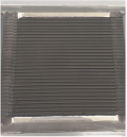}
    \caption{}
    \includegraphics[width=1\textwidth, height = 4.2cm]{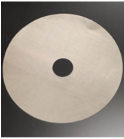}
    \caption{}
\end{subfigure}
\begin{subfigure}[t]{0.45\columnwidth}
    \includegraphics[width=1\textwidth, height = 4.2cm]{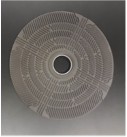}
    \caption{}
    \includegraphics[width=1\textwidth, height = 4.2cm]{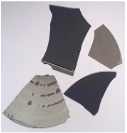}
    \caption{}
\end{subfigure}
\caption{Dataset samples: (A) HTEL – (B) Ni-Mesh – (C) Steel – (D) Samples from mixed origins and states.}
\label{fig:samples}
\end{figure}

Our samples represent shredded pieces from electrolyzer cells of three major materials: High Temperature electrolyzers (HTEL) ceramics, Ni-mesh, and interconnector steel plates \cite{fleischhauer2015strength}, originally from two different sources in two states: new samples and old end-of-life samples. Fig. \ref{fig:samples} shows the three main components of HTEL cells and their different life-time states. It is worth noting that Ni-Mesh has an identical color to the Steel plate but a different and finer surface texture that causes high light reflection. HTEL ceramics and interconnector steel have two different functional faces, creating in total five classes of interests: Mesh, Steel black, Steel gray, HTEL Cathode, and HTEL Anode. These classes represent the expected material surfaces exposed to imaging sensors in electrolyzers recycling streams. 

\begin{figure}[!ht]
\centering
\begin{subfigure}[t]{0.32\columnwidth}
    \makebox[0pt][r]{\makebox[30pt]{\raisebox{25pt}{\rotatebox[origin=c]{90}{\small Sample 16}}}}%
    \includegraphics[width=1\textwidth]{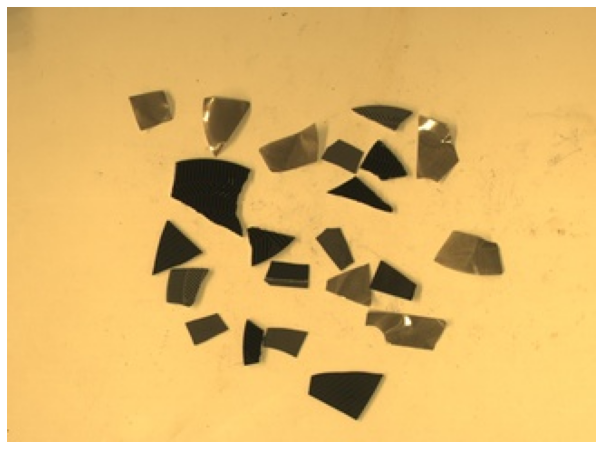}
    \makebox[0pt][r]{\makebox[30pt]{\raisebox{25pt}{\rotatebox[origin=c]{90}{\small Sample 29}}}}%
    \includegraphics[width=\textwidth]{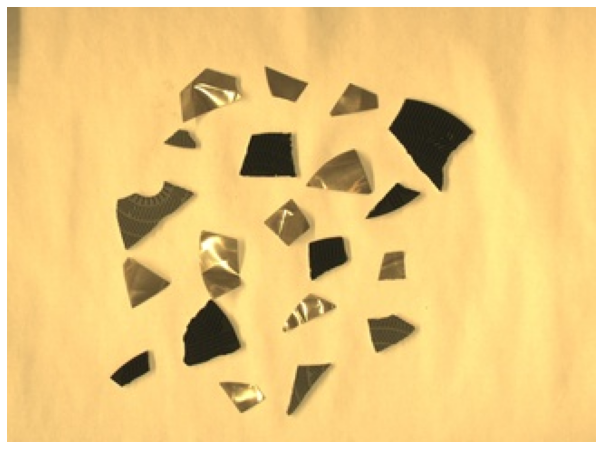}
    \makebox[0pt][r]{\makebox[30pt]{\raisebox{25pt}{\rotatebox[origin=c]{90}{\small Sample 54}}}}%
    \includegraphics[width=\textwidth]{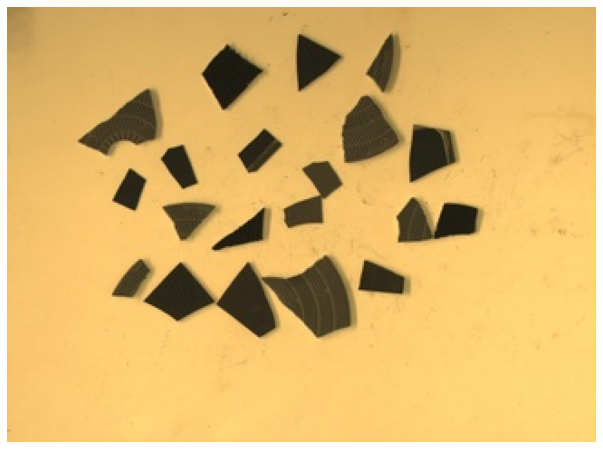}
    \caption{RGB}
\end{subfigure}
\begin{subfigure}[t]{0.32\columnwidth}
    \includegraphics[width=1\textwidth]{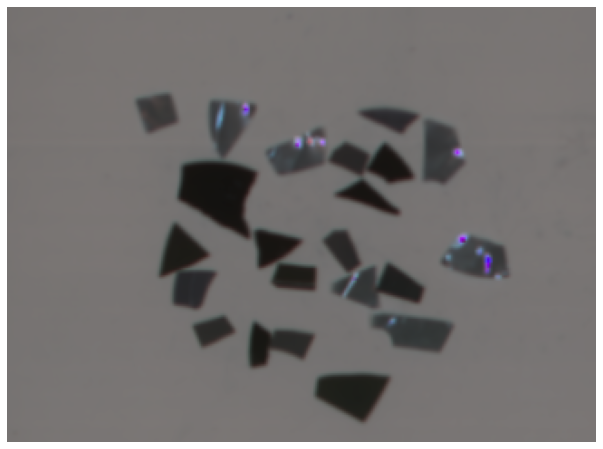}
    \includegraphics[width=\textwidth]{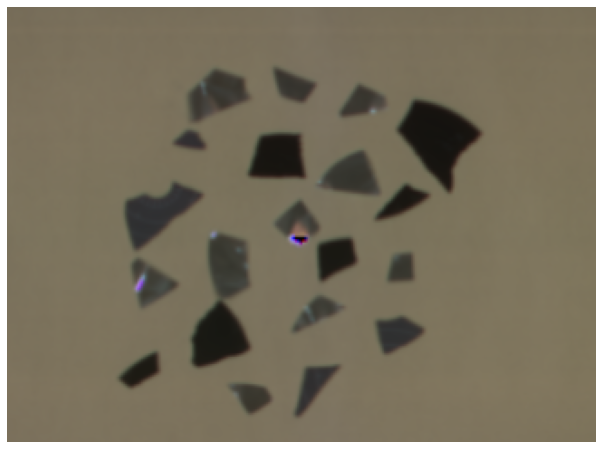}
    \includegraphics[width=\textwidth]{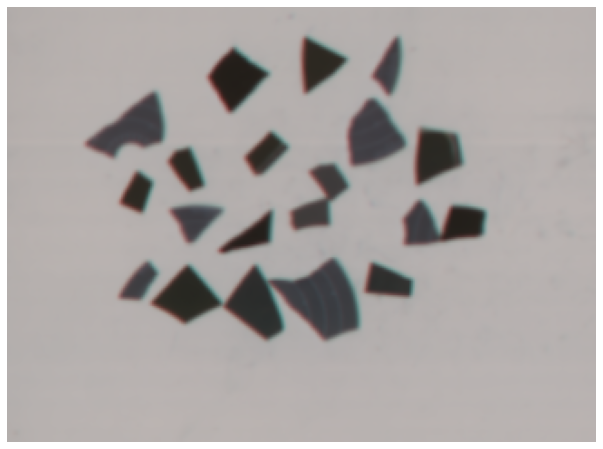}
    \caption{HSI}
\end{subfigure}
\begin{subfigure}[t]{0.32\columnwidth}
    \includegraphics[width=1\textwidth]{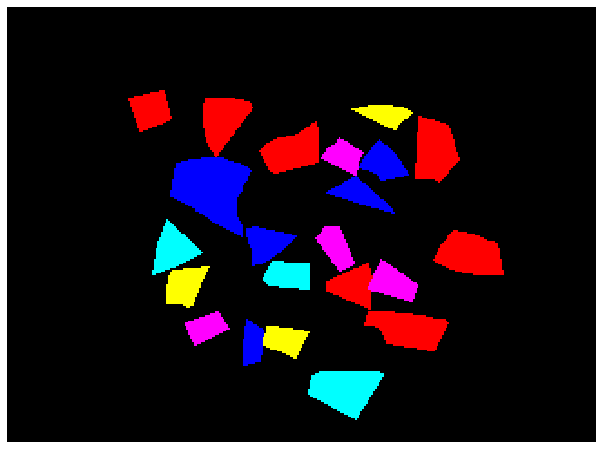}
    \includegraphics[width=\textwidth]{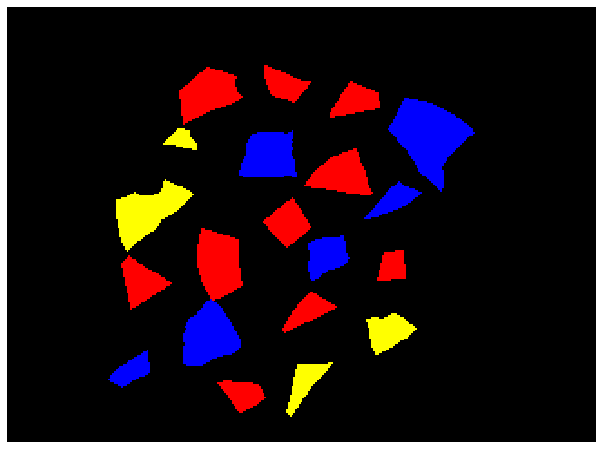}
    \includegraphics[width=\textwidth]{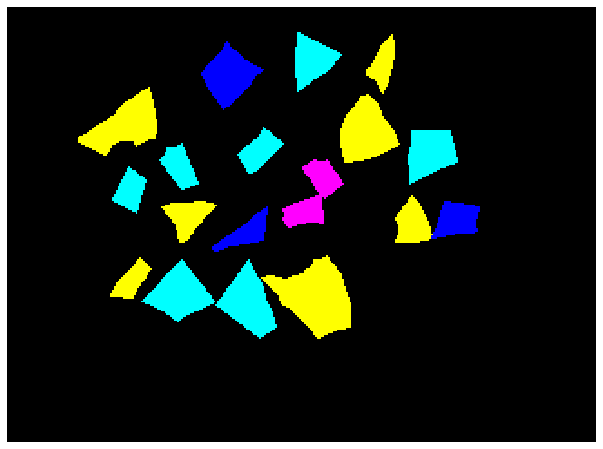}
    \caption{Ground Truth}
\end{subfigure}
\scriptsize
    \colorbox{Class_1!}{Mesh}
    \colorbox{Class_2!}{Steel-Black}
    \colorbox{Class_3!}{Steel-Gray}
    \colorbox{Class_4!}{HTEL-Anode}
    \colorbox{Class_5!}{HTEL-Cathode}
\caption{The triplet images per scan consist of the high spatial resolution RGB image, the HSI data cube (false color representation, using bands 2069 nm, 1792 nm, and 1401 nm), and the ground truth.}
\label{fig:datset_samples}
\end{figure}

The samples of HTEL stack cells were physically shredded to simulate real-world recycling conditions, and the resulting material fragments were then systematically scanned as seen in Fig. \ref{fig:samples} (D). Scans containing different numbers of classes were generated: 6 scans containing only a single class, 31 scans containing two classes, 9 scans with three classes, 4 scans with four classes, and 5 scans with five classes. This structured approach enables both controlled single-class learning and multi-class classification. 
For each scan of the samples, two scans were performed, one for each side of the material surface (front and back), ensuring comprehensive spectral coverage for all targeted classes. 

The scanning device is AisaFENIX (Spectral Imaging Ltd) push-broom HSI camera with 450 bands in the VNIR to SWIR wavelength range ( [400-2500] nm, spectral sampling VNIR: 3,4 nm, SWIR: 5,7 nm, spatial resolution: 384 pixels/line). High-resolution spatial data for geometric information was acquired in parallel using an LT-400 CL 3 CMOS RGB line scan camera (spatial resolution: 4096 pixels).

The final dataset contains a total of 55 triplets of images with spatial size $240\times325$ consisting of co-registered high-resolution RGB images and their high spectral resolution HSI twins, in addition to the ground truth masks. A total of 424,169 labeled pixels out of 4,290,000 pixel vectors are calculated from the ground truth masks. Fig. \ref{fig:datset_samples} visualizes scans 16, 29, and 54  consisting of the RGB images, false color representation of the HSI [2069, 1792, 1401] nm, and the ground truth masks. The RGB image provides high-resolution spatial features that support classification. However, since the samples' shape can differ depending on the end-of-life status, spectral features are the main classification key.

In order to obtain efficient data processing, the first 50 and last 40 bands were discarded from the HSI to eliminate noisy acquisitions, resulting in HSI with 360 bands. Then, spectral binning was applied by averaging every two adjacent bands in each HSI into one band. This reduces the input spectral dimension to 180 bands and mitigates information abundance without sacrificing spectral characterization, leading to better convergence.

\begin{figure}[t]
\centering
\includegraphics[width=0.99\columnwidth]{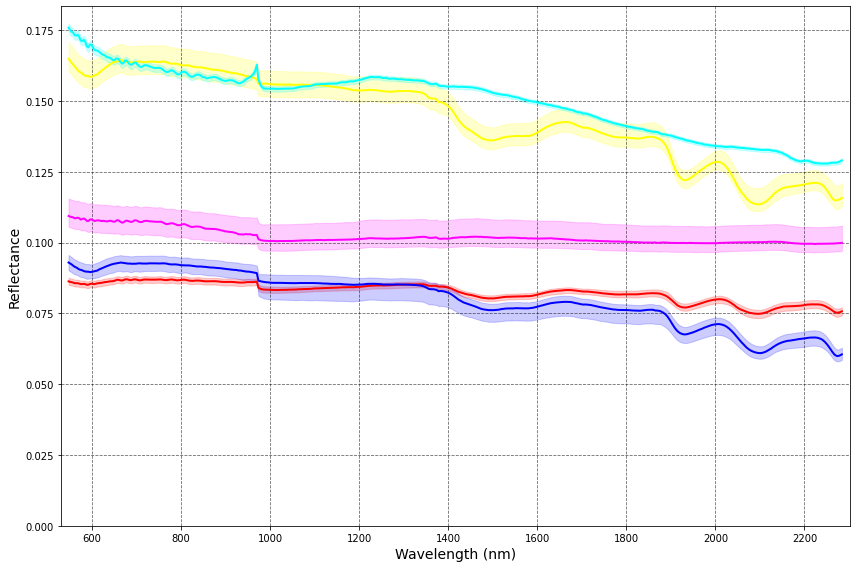}
\scriptsize
    \colorbox{Class_1!}{Mesh}
    \colorbox{Class_2!}{Steel-Black}
    \colorbox{Class_3!}{Steel-Gray}
    \colorbox{Class_4!}{HTEL-Anode}
    \colorbox{Class_5!}{HTEL-Cathode}
\caption{The spectral signatures of the five classes}
\label{signature}
\end{figure}

Fig. \ref{signature} shows the spectral signatures of the five classes that act as the input of the further processing models. The overlap in the spectral profiles of different classes can be observed. Moreover, the spectra of the metal material types are flat and invariant, and the dark surface of two classes (Steel black and Mesh) from different components causes high absorbance in the VNIR-SWIR, resulting in very low reflectance signals.
\begin{figure}[t]
\centering
\includegraphics[width=0.99\columnwidth]{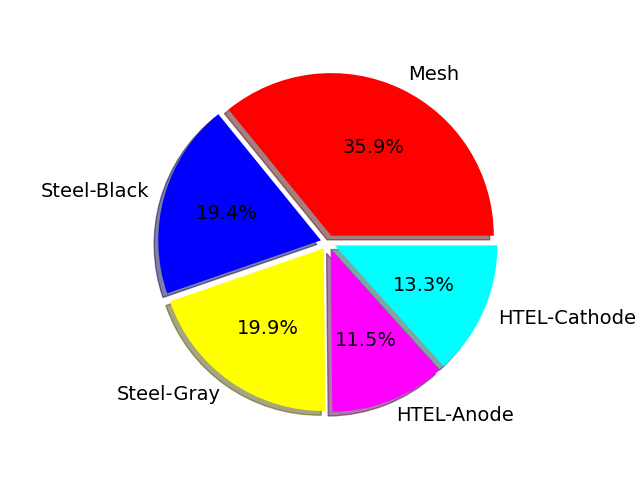}
\caption{The class distribution of the training set.}
\label{stats_train}
\end{figure}

\begin{figure}[b]
\centering
\includegraphics[width=1.1\columnwidth]{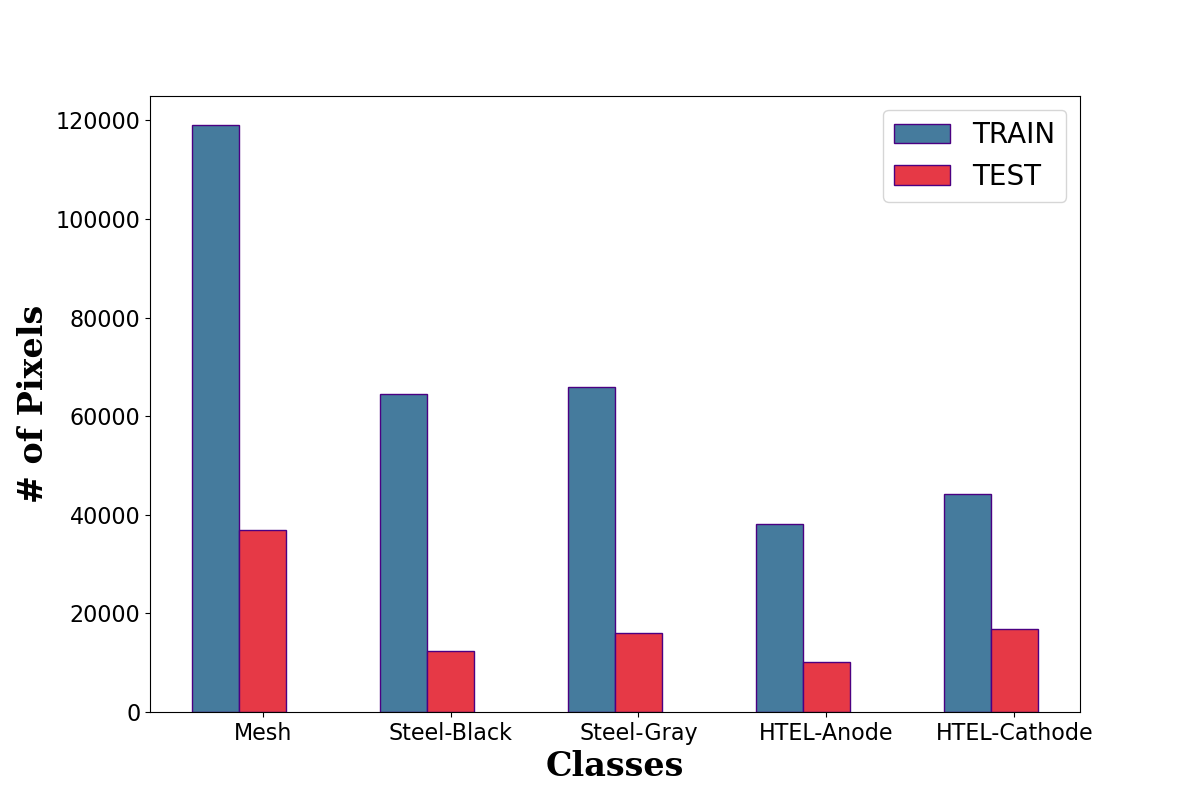}
\caption{A bar plot comparing the number of pixels between the training set and test set per class.}
\label{dataset-bar}
\end{figure}

\subsection{Statistics}

From the total dataset, 44 images were used for training and 11 for testing, covering diverse class combinations and acquisition conditions. In total, 336,215 pixels were used for training and 87,954 for testing.

Fig. \ref{stats_train} illustrates the class distribution of the training samples, while Fig. \ref{dataset-bar} presents the relative sizes of the training (blue) and test (red) sets across the five classes. As shown, the dataset is notably imbalanced, with the "Mesh" class comprising 35.9\% of the training samples, substantially more than the other classes, that are more evenly distributed. If unaddressed, such a class imbalance can lead to biased models with poor generalization on underrepresented classes. To mitigate this issue, we applied a weighted cross-entropy loss, where class weights are computed inversely proportional to their frequencies in the training set. This approach increases the influence of minority classes during parameters optimization, encouraging the model to learn more representative and invariant features across all training data.

\section{Experiments and Analysis}
\label{sec:experiements}

In this section, we present the processing workflow, in addition to the evaluated models, highlighting key differences in their application on RGB images and HSI data with the input configurations.
From Figure Fig.\ref{acquisition}, following the RGB and HSI data acquisition, we preprocess each modality in parallel; HSI preprocessing involves reflectance conversion and normalization, while RGB preprocessing includes applying the pre-trained Segment Anything Model (SAM) \cite{kirillov2023segment} and Grounding Dino \cite{liu2024grounding} using the method in \cite{arbash2023masking} to segment foreground objects from the background, obtaining the objects instance segmentation masks simultaneously. These masks are then used to generate masked HSI data cubes, enabling processing models to focus exclusively on object materials pixels only. The preprocessed data is processed with the trained ML and the Transformer-based modalities for pixel-wise electrolyzers classification. Finally, we overlay the instance masks on the pixel-wise classification maps, and object-wise classification is achieved through majority voting within object polygons defined by the instance masks generated by the zero-shot models.

\subsection{SOTA Models}

We investigated the application of electrolyzers classification using the provided dataset using several representative ML and SOTA Transformer-based \cite{vaswani2017attention} HSI classification models. Transformers became the backbone of all SOTA data processing models due to their core computation process, i.e., self-attention, which serves as a suitable mechanism for detecting the spectral features when the data is properly tokenized.
In the spectral domain, spectral fingerprints represent unique changes in the spectral signature at specific wavelengths, which characterize the light absorption features of the material. Spectral features are encoded in the HSI data using two variables: the reflectance value and its wavelength position in the spectrum. To effectively detect these patterns, a computational framework is required that can model the relationships and affinities between input tokens that together represent these two dimensions. The self-attention mechanism within Transformer architectures enables this by dynamically updating the representation of each token based on contextual information from all other tokens. This makes Transformers particularly suitable for extracting and modeling complex spectral features in HSI data.
The selected Transformer models were among the first to be applied to images, including RGB and HSI, making them suitable for exposing fundamental challenges and limitations of the modality compared to more recent, specialized variants. 

The evaluated models are: 
\begin{itemize}

\item Traditional machine learning (ML) models, including Random Forest, K-Nearest Neighbors (KNN), and Support Vector Machine (SVM) were employed to process individual hyperspectral vectors.
Given the computational demands associated with processing large volumes of high-dimensional HSI data, these models were implemented using Dask, a parallel computing library. By leveraging Dask arrays and pipelines with a chunk size of 10,000, we efficiently distributed the computational workload, significantly reducing both memory usage and processing time.

\item \textbf{Vision Transformer} (ViT) is the Transformer-encoder architecture that introduced the adaptation of Transformers \cite{NIPS2017_3f5ee243} to image classification tasks \cite{dosovitskiy2021an}. In its original implementation, an input image is divided into non-overlapping 16×16 patches along the spatial dimensions of the RGB image. For deploying ViT on HSI, we extract spectral patches from the same spatial location to predict the class label of the center pixel. Each patch from a different band is then linearly embedded and treated as a token. The model applies a self-attention mechanism to iteratively update these token representations by incorporating contextual information from all other tokens, enabling the network to capture global dependencies and effectively perform scene understanding and image classification.
\item \textbf{SpectralFormer} \cite{hong2021spectralformer} was one of the first models to adapt Transformer architectures specifically for HSI classification. SpectralFormer is built on the ViT framework, on top of which two architectural features tailored for HSI processing are introduced: i) groupwise spectral embedding (GSE), which enriches patch embeddings by emphasizing spectral features, ii) cross-layer adaptive fusion (CAF) to enhance the feature integration across encoder layers. SpectralFormer is implemented in two configurations: a \textbf{pixel-wise} version that classifies using only the center pixel's spectral vector, and a \textbf{patch-wise} version that processes a full 9×9 spatial patch (81 vectors), incorporating both spectral and local spatial context for improved classification accuracy. A patch size of $9\times9$ was selected to balance memory efficiency with classification accuracy, while also minimizing the risk of mixed spectral signatures. Larger patches tend to include pixels from multiple classes, causing more ambiguity, thus reducing the model's performance.

\item \textbf{Multimodal Fusion Transformer (MFT)} \cite{roy2023multimodal}, following the SpectralFormer adaptation of Transformer-based encoders on HSI data, MFT is a ViT-based neural network designed for pixel-wise classification of HSI, with architectural enhancements to integrate a second modality. Built upon a standard Vision Transformer encoder, MFT introduces modifications that allow the encoder to process another modality via the classification token (CLS). Initially, both modalities undergo feature extraction via new CNN blocks. The primary modality, HSI, is processed through a combination of 2D and 3D convolutional layers to capture spatial-spectral relationships, then is tokenized, while the secondary modality, RGB in our case, is processed through a separate CNN pathway. The RGB-derived features are then incorporated into the Transformer via the CLS token, enabling cross-modal interaction. MFT employs cross-modality attention mechanisms to facilitate fusion between HSI and RGB features. As with SpectralFormer, tokens are generated from the spectral bands at the same spatial location, preserving the integrity of pixel-wise classification while enabling multimodal learning \cite{roy2023multimodal}. 
    
\end{itemize}

Moreover, to improve model generalization, we applied a combination of eight spectral and spatial augmentation techniques during training. Spectral augmentations included band shifting, spectral smoothing, noise addition, scaling, and channel dropping, while spatial augmentations consisted of image rotation, translation, and flipping. These transformations were applied on the fly during training, effectively increasing the size of the dataset by a factor of eight. This dynamic augmentation strategy enabled training over 600 epochs with a reduced learning rate of 1e{-7}. For SpectralFormer, we adopted a groupwise spectral embedding (GSE) size of 7, consistent with the original implementation. All models were trained using the Adam optimizer with a mini-batch size of 512. 

\subsection{Pixel-wise Classification Evaluations}

\begin{table*}[th]
    \centering
    \footnotesize
        \caption{Pixel-wise classification results for the different models in terms of the F1 score per class, overall accuracy (OA), and average accuracy (AA).}
    \label{tab:pixelwise}
    \begin{tabular}{c||c||c|c||c||c|c|c}
    \toprule
    \multirow{2}{*}{\textbf{Classes}} & \multicolumn{1}{c||}{\textbf{Multimodality}} & \multicolumn{2}{c||}{\textbf{SpectralFormer}} & \multicolumn{1}{c||}{\textbf{Transformers}} & \multicolumn{3}{c}{\textbf{Conventional Classifiers}} \\
    \cline{2-2} \cline{3-4} \cline{5-5} \cline{6-8}
    & \textbf{MFT 9x9} & \textbf{Patch-wise 9x9} & \textbf{Pixel-wise} & \textbf{ViT} & \textbf{SVM} & \textbf{RF} & \textbf{KNN} \\
    \midrule
    1 (Mesh) & \textbf{86.03} & 74.51 & 78.88 & 75.88 & 80.62 & 52.19 & 45.29 \\
    2 (Steel - Cathode)& \textbf{72.27} & 31.39 & 54.67 & 27.42 & 3.26 & 7.41 & 21.67 \\
    3 (Steel - Anode)& \textbf{75.90} & 32.78 & 52.06 & 45.77 & 15.20 & 14.42 & 6.49 \\
    4 (HTEL - Anode)& \textbf{40.68} & 25.77 & 36.47 & 31.74 & 28.89 & 1.05 & 3.13 \\
    5 (HTEL - Cathode)& \textbf{36.04} & 34.37 & 30.71 & 38.48 & 3.07 & 3.24 & 16.72 \\
    \midrule
    OA (\%) & \textbf{69.30} & 47.92 & 59.78 & 51.81 & 47.08 & 35.17 & 25.28 \\
    AA (\%) & \textbf{64.29} & 39.95 & 51.15 & 43.81 & 34.98 & 19.68 & 22.14 \\
    \bottomrule
    \end{tabular}
\end{table*}

\begin{figure}[th]
    \centering
    \begin{adjustbox}{width=\columnwidth}
        \begin{tabular}{c c c c}
            & \textbf{\Huge Sample 1} & \textbf{\Huge Sample 42} & \textbf{\Huge Sample 47}\\ 
            
            \makebox{\raisebox{80pt}{\rotatebox[origin=c]{90}{\textbf{\Huge Ground Truth}}}} & 
            \includegraphics[width=\columnwidth]{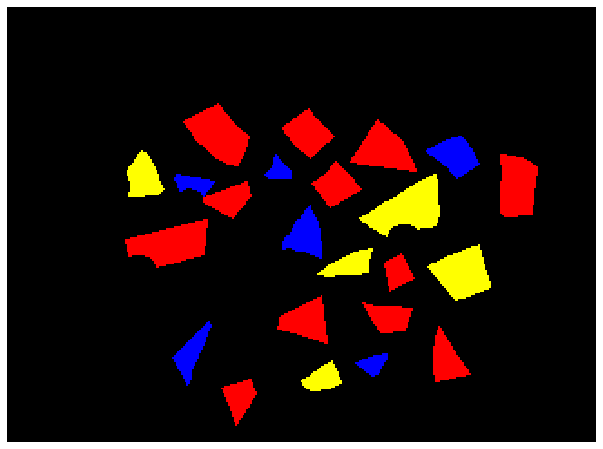} & 
            \includegraphics[width=\columnwidth]{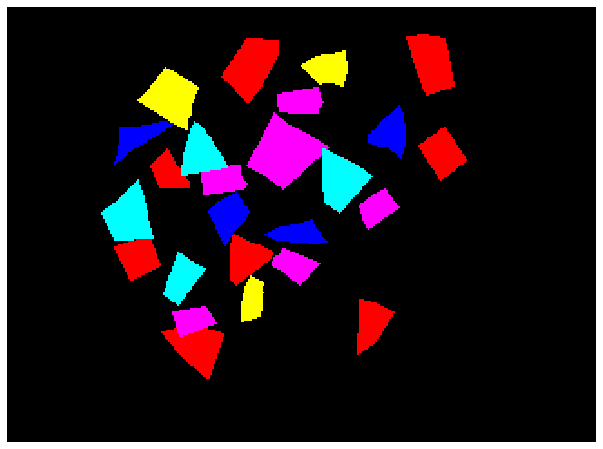} & 
            \includegraphics[width=\columnwidth]{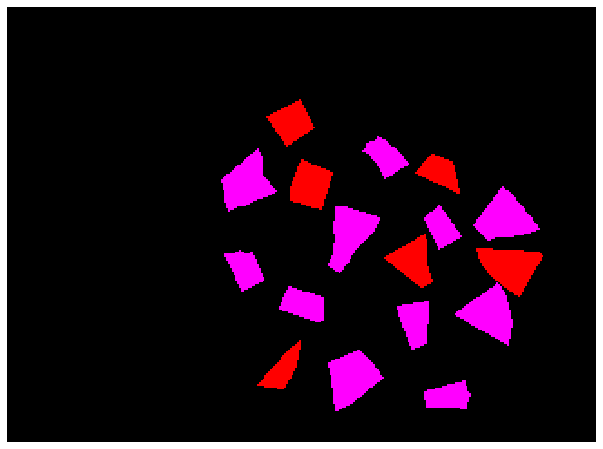} \\

            \makebox{\raisebox{80pt}{\rotatebox[origin=c]{90}{\textbf{\Huge MFT}}}} & 
            \includegraphics[width=\columnwidth]{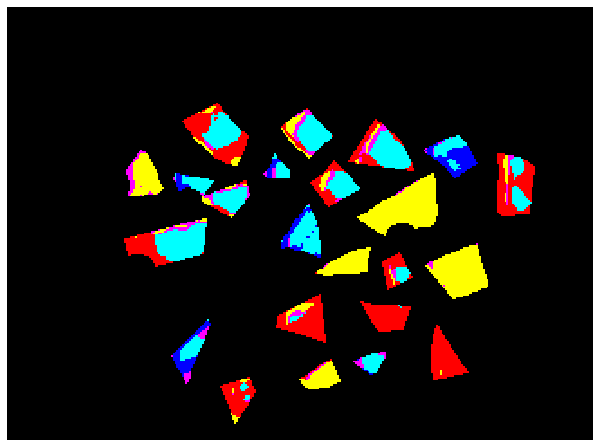} & 
            \includegraphics[width=\columnwidth]{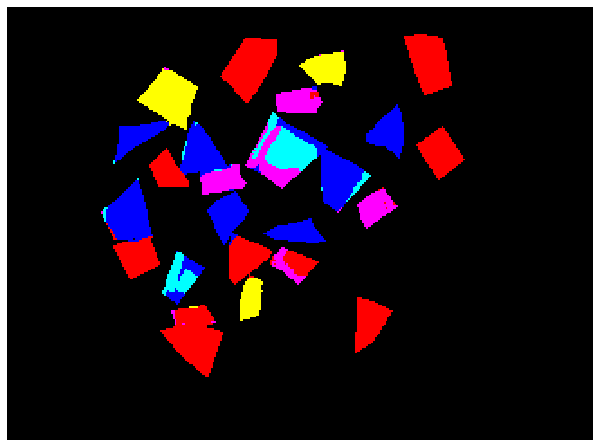} & 
            \includegraphics[width=\columnwidth]{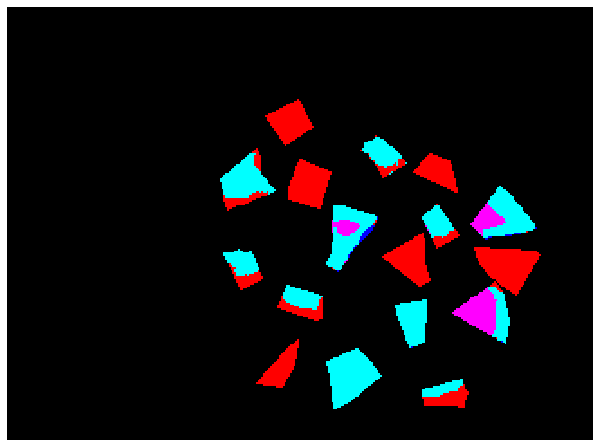} \\

            \makebox{\raisebox{80pt}{\rotatebox[origin=c]{90}{\textbf{\Huge SF-Patch}}}} & 
            \includegraphics[width=\columnwidth]{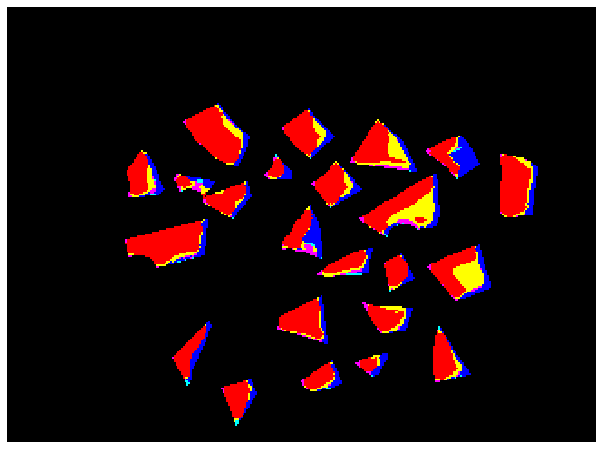} & 
            \includegraphics[width=\columnwidth]{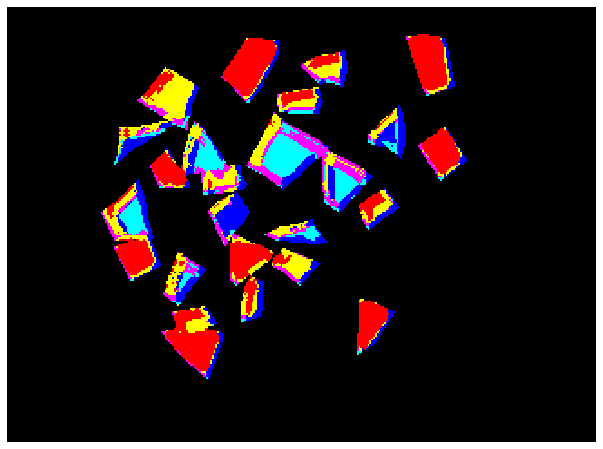} & 
            \includegraphics[width=\columnwidth]{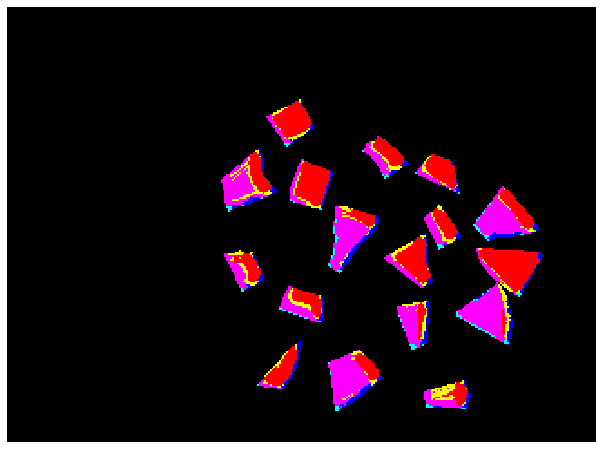} \\

            \makebox{\raisebox{80pt}{\rotatebox[origin=c]{90}{\textbf{\Huge SF-Pixel}}}} & 
            \includegraphics[width=\columnwidth]{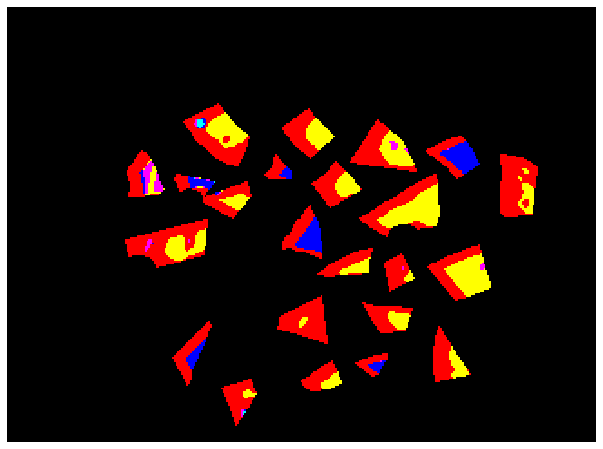} & 
            \includegraphics[width=\columnwidth]{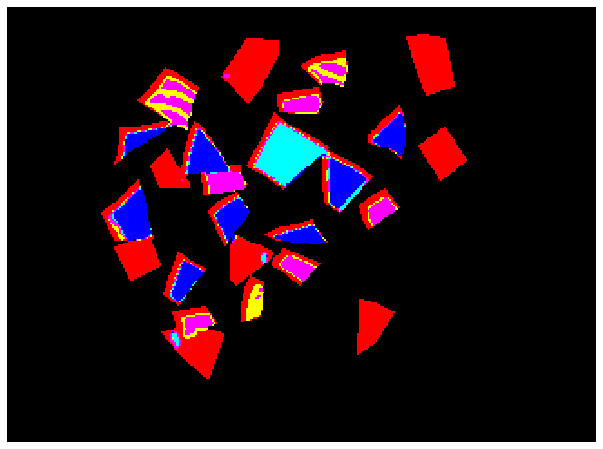} & 
            \includegraphics[width=\columnwidth]{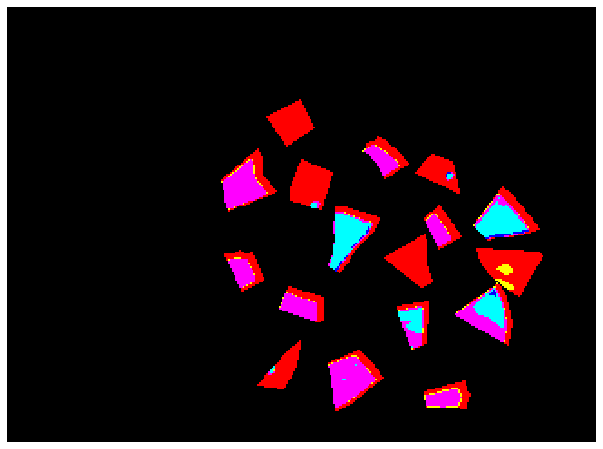} \\

            \makebox{\raisebox{80pt}{\rotatebox[origin=c]{90}{\textbf{\Huge ViT}}}} & 
            \includegraphics[width=\columnwidth]{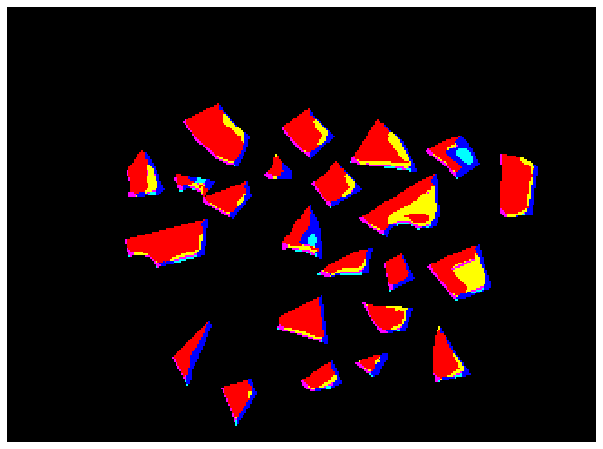} & 
            \includegraphics[width=\columnwidth]{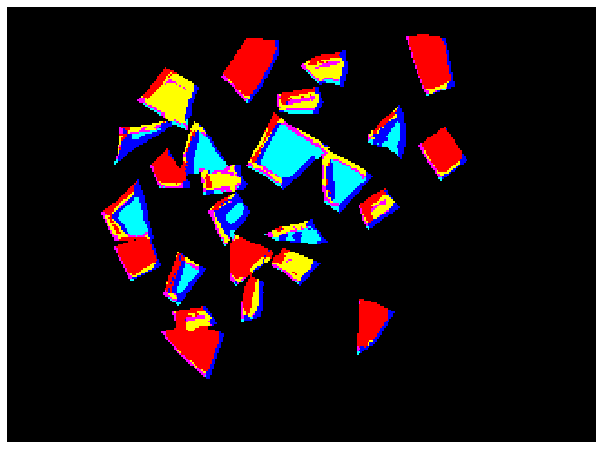} & 
            \includegraphics[width=\columnwidth]{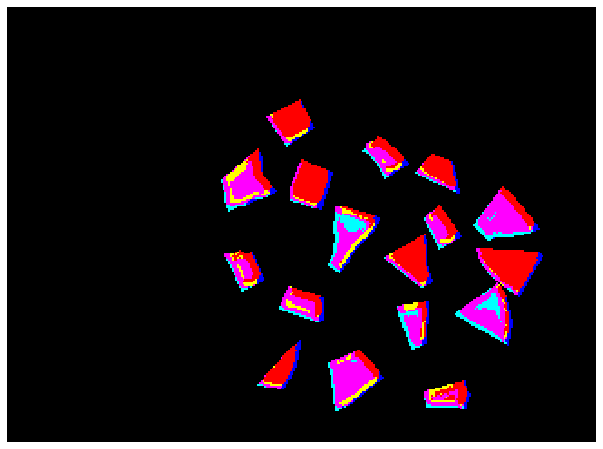} \\

            \makebox{\raisebox{80pt}{\rotatebox[origin=c]{90}{\textbf{\Huge SVM}}}} & 
            \includegraphics[width=\columnwidth]{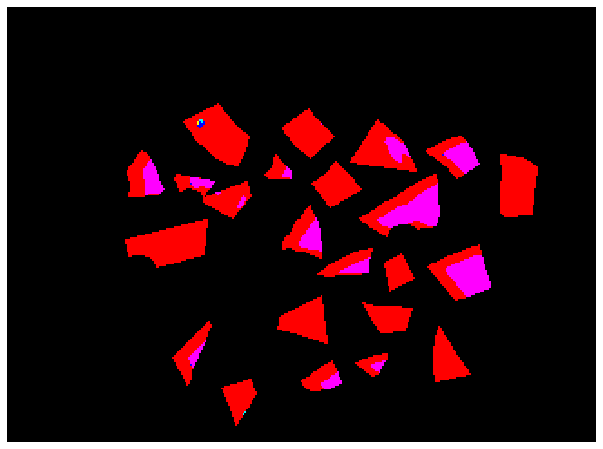} & 
            \includegraphics[width=\columnwidth]{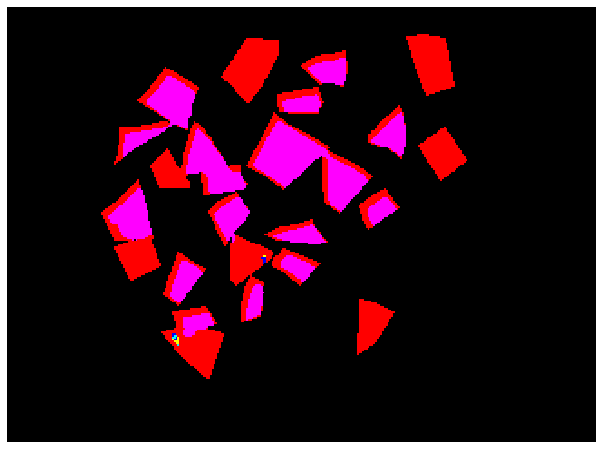} & 
            \includegraphics[width=\columnwidth]{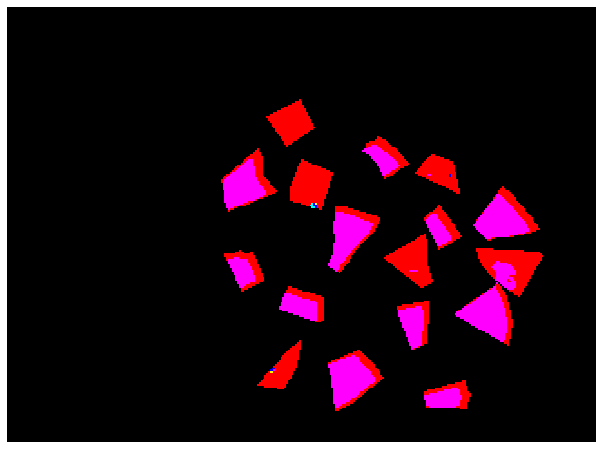} \\

            \makebox{\raisebox{80pt}{\rotatebox[origin=c]{90}{\textbf{\Huge RF}}}} & 
            \includegraphics[width=\columnwidth]{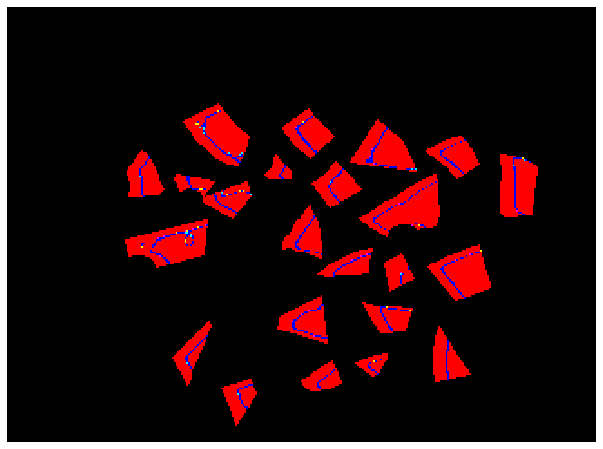} & 
            \includegraphics[width=\columnwidth]{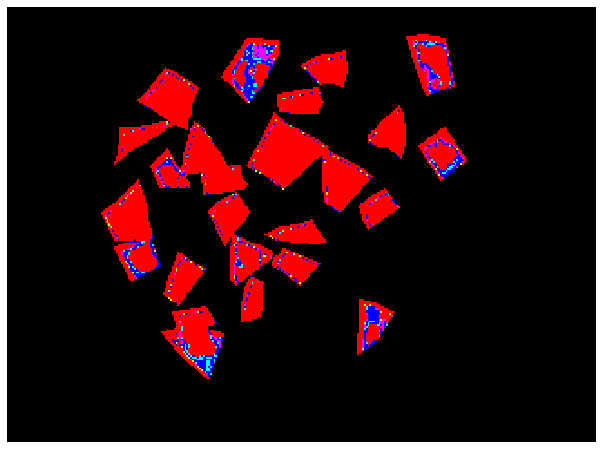} & 
            \includegraphics[width=\columnwidth]{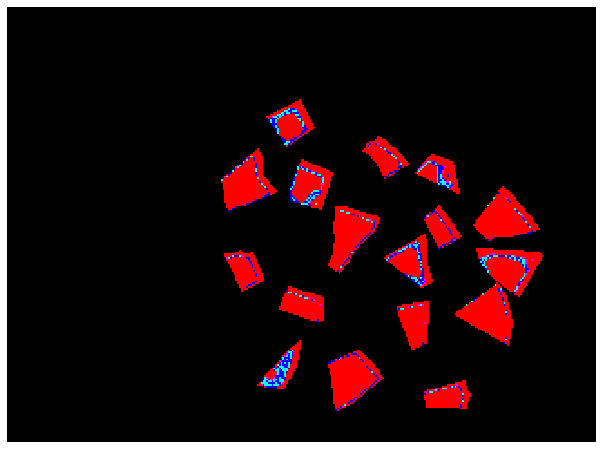} \\

            \makebox{\raisebox{80pt}{\rotatebox[origin=c]{90}{\textbf{\Huge KNN}}}} & 
            \includegraphics[width=\columnwidth]{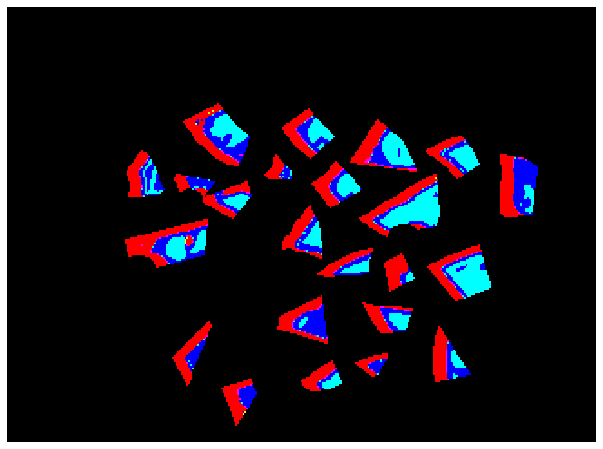} & 
            \includegraphics[width=\columnwidth]{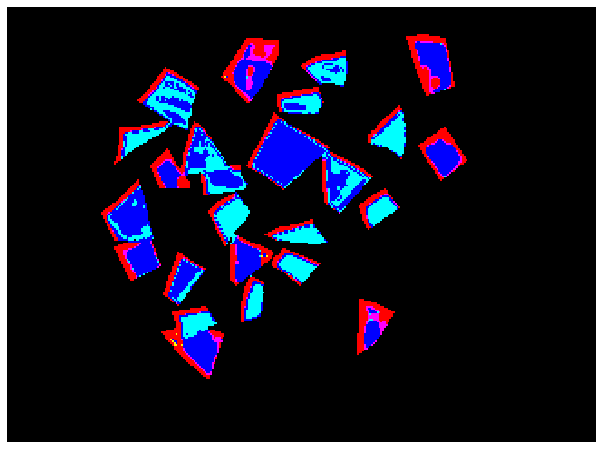} & 
            \includegraphics[width=\columnwidth]{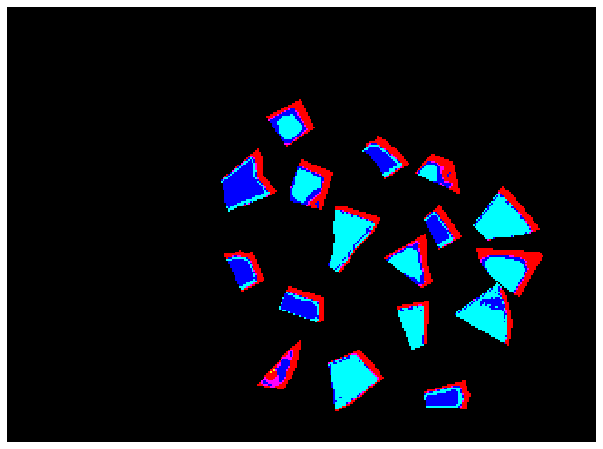} \\
        \end{tabular}
    \end{adjustbox}\\
    \scriptsize
    \colorbox{Class_1!}{Mesh}
    \colorbox{Class_2!}{Steel-Black}
    \colorbox{Class_3!}{Steel-Gray}
    \colorbox{Class_4!}{HTEL-Anode}
    \colorbox{Class_5!}{HTEL-Cathode}
    \caption{Pixel-wise classification prediction maps on samples 1, 42, and 47 from the test set.}
    \label{fig:pixewise predictions}
\end{figure}

HSI pixel-wise classification results offer valuable insights into model behavior and performance, highlighting areas for potential optimization. Fig. \ref{fig:pixewise predictions} visualizes the pixel-wise prediction maps of the deployed models, and Table \ref{tab:pixelwise} presents the pixel-level classification performance of all evaluated models, reporting per-class F1 scores along with overall accuracy (OA) and average accuracy (AA). The results are organized from left to right with the MFT multimodality model, followed by the single-modality ones, SpectralFormer and ViT, and ending with the classical machine learning baselines. The table indicates the following: 

\begin{itemize}
    \item The superior performance of the MFT model — which exploits both RGB and HSI modalities—is evident, achieving the highest scores across all evaluation metrics. This outcome is in line with expectations, as the high spatial resolution of the RGB images contributes significantly to class discrimination. These results highlight the power of multimodal approaches and underline the benefits of integrating complementary spatial and spectral features for more robust classification. 
    \item The pixel-wise SpectralFormer performs second best after MFT, outperforming both the patch-wise SpectralFormer and the ViT model. This distinction provides important insights into the behavior of Transformer-based encoders when applied to HSI data. While GSE in SpectralFormer \cite{hong2021spectralformer} enriches token representations by integrating information from neighboring spectral bands, emphasizing spectral feature learning, it also introduces convergence issues when the input patch contains mixed multi-class spectra, a problem illustrated in Fig. \ref{fig:patch}.
As a result, the patch-wise SpectralFormer suffers from a performance drop compared to its pixel-wise counterpart. The pixel-wise variant avoids this issue by processing individual spectral vectors, allowing the Transformer to focus exclusively on the spectral features of a single material without impurities from neighboring classes. While this comes at the cost of spatial context, it guarantees spectral consistency within the token. To restore spatial awareness without compromising spectral purity, object-level context is later recovered via independent zero-shot modalities and post-processed via object-guided majority voting. This strategy provides a balance between spectral precision and spatial reasoning, ultimately improving classification robustness.
    \item In contrast, the ViT does not include the GSE or CAF modules from SpectralFormer. Instead, it tokenizes each spectral channel within an HSI patch as an independent input token. As a result, mixed-material signatures affect less when patches contain spectra from multiple classes than when patches are processed with GSE. The comparable performance between ViT and the pixel-wise SpectralFormer highlights an important drawback: while including more than one spectral vector (e.g., using the entire patch instead of just the center pixel) can improve classification by introducing richer contextual information, it also increases the risk of including mixed spectral signatures in the input potentially leading to confusion during learning.
    \item The performance of the classical ML models is heavily biased toward the "Mesh" class, with significantly lower accuracy in detecting the remaining classes. This imbalance becomes even more apparent when object-wise classification is applied to the pixel-level predictions, further demonstrating the limited generalization of the models across different material types.

\end{itemize}

\subsection{Object-wise Classification Evaluations}

\begin{table*}[h]
    \centering
    \footnotesize
        \caption{Object-wise classification results via majority voting 
         for the different models in terms of the F1 score per class, overall accuracy (OA), and average
accuracy (AA).}
    \label{tab:majority}
    \begin{tabular}{c||c||c|c||c||c|c|c}
    \toprule
    \multirow{2}{*}{\textbf{Classes}} & \multicolumn{1}{c||}{\textbf{Multimodality}} & \multicolumn{2}{c||}{\textbf{SpectralFormer}} & \multicolumn{1}{c||}{\textbf{Transformers}} & \multicolumn{3}{c}{\textbf{Conventional Classifiers}} \\
    \cline{2-2} \cline{3-4} \cline{5-5} \cline{6-8}
    & \textbf{MFT 9x9} & \textbf{Patch-wise 9x9} & \textbf{Pixel-wise} & \textbf{ViT} & \textbf{SVM} & \textbf{RF} & \textbf{KNN} \\
    \midrule
    1 (Mesh) & \textbf{99.93} & 99.91 & 99.92 & 99.91 & 99.92 & 99.92 & 99.93 \\
    2 (Steel - Black)& 82.65 & 76.92 & 82.26 & \textbf{83.00} & 81.59 & 64.71 & 51.63 \\
    3 (Steel - Gray)& \textbf{80.35} & 41.55 & 73.84 & 58.85 & 1.24 & 3.24 & 27.92 \\
    4 (HTEL - Anode)& \textbf{70.19} & 47.39 & 63.15 & 64.81 & 4.81 & 4.13 & 1.37 \\
    5 (HTEL - Cathode)& 39.54 & 33.83 & 45.41 & \textbf{51.40} & 40.17 & 0.16 & 0.68 \\
    \midrule
    OA (\%) & \textbf{98.16} & 97.36 & 98.02 & 98.10 & 96.31 & 95.70 & 95.77 \\
    AA (\%) & \textbf{74.55} & 58.84 & 70.35 & 68.32 & 54.77 & 40.31 & 43.03 \\
    \bottomrule
    \end{tabular}
\end{table*}

\begin{figure*}[!ht]
\centering
\begin{subfigure}[t]{0.19\textwidth}
    \makebox[0pt][r]{\makebox[30pt]{\raisebox{25pt}{\rotatebox[origin=c]{90}{Sample 1}}}}%
    \includegraphics[width=\textwidth]{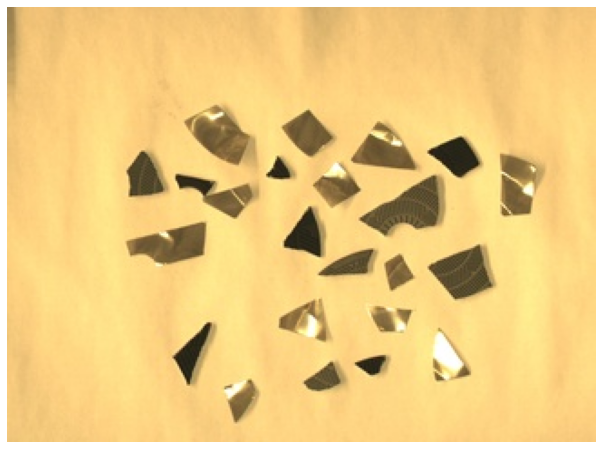}
    \makebox[0pt][r]{\makebox[30pt]{\raisebox{25pt}{\rotatebox[origin=c]{90}{Sample 42}}}}%
    \includegraphics[width=\textwidth]{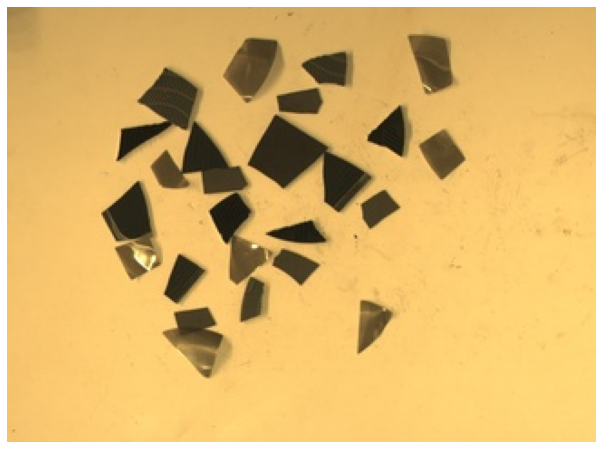}
    \caption{Input}
\end{subfigure}
\begin{subfigure}[t]{0.19\textwidth}
    \includegraphics[width=\textwidth]{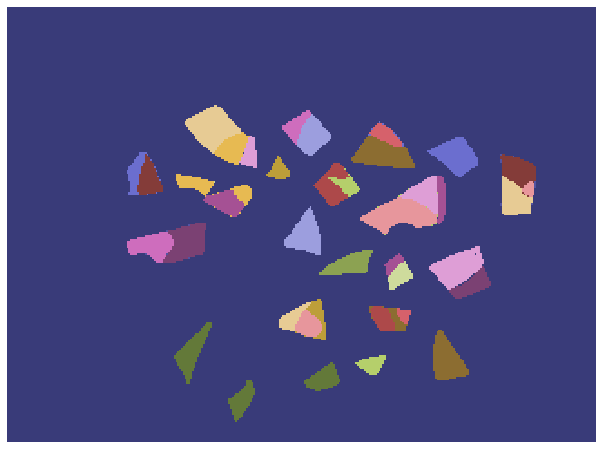}
    \includegraphics[width=\textwidth]{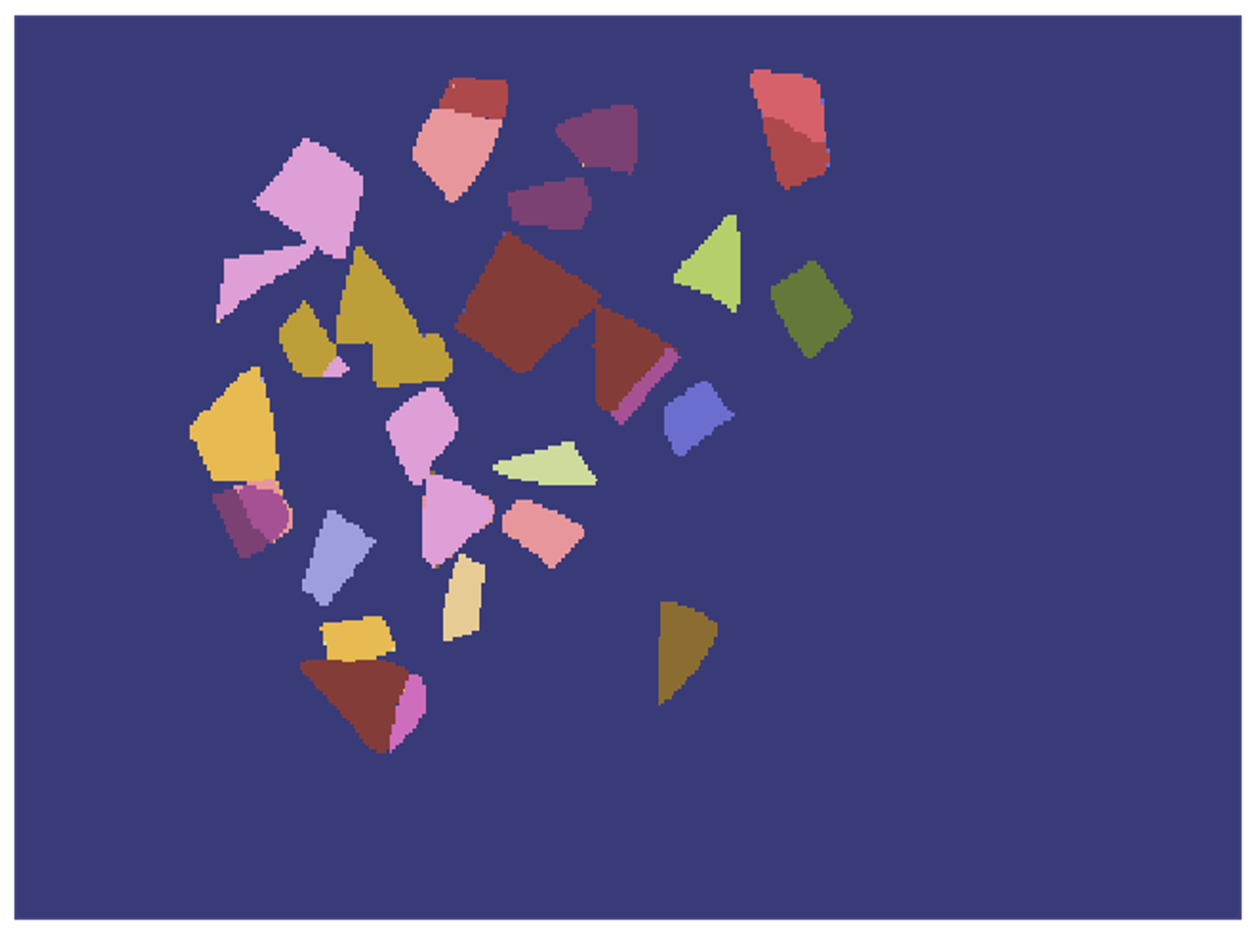}
    \caption{Instance Segmentation}
\end{subfigure}
\begin{subfigure}[t]{0.19\textwidth}
    \includegraphics[width=\textwidth]{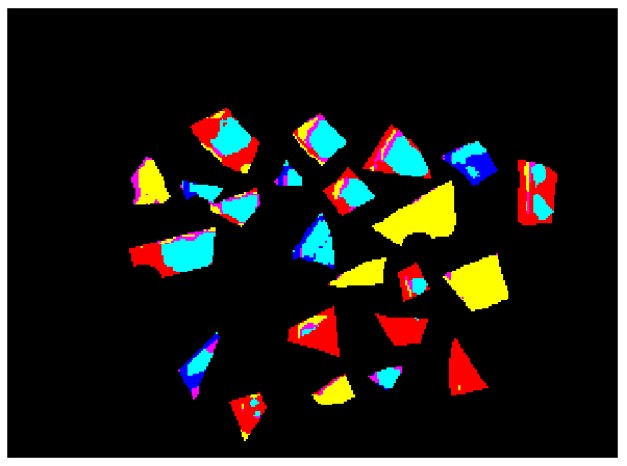}
    \includegraphics[width=\textwidth]{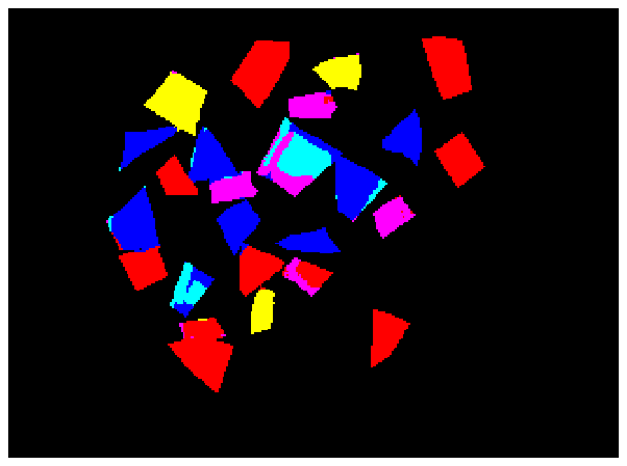}
    \caption{Pixel classification}
\end{subfigure}
\begin{subfigure}[t]{0.19\textwidth}
    \includegraphics[width=\textwidth]{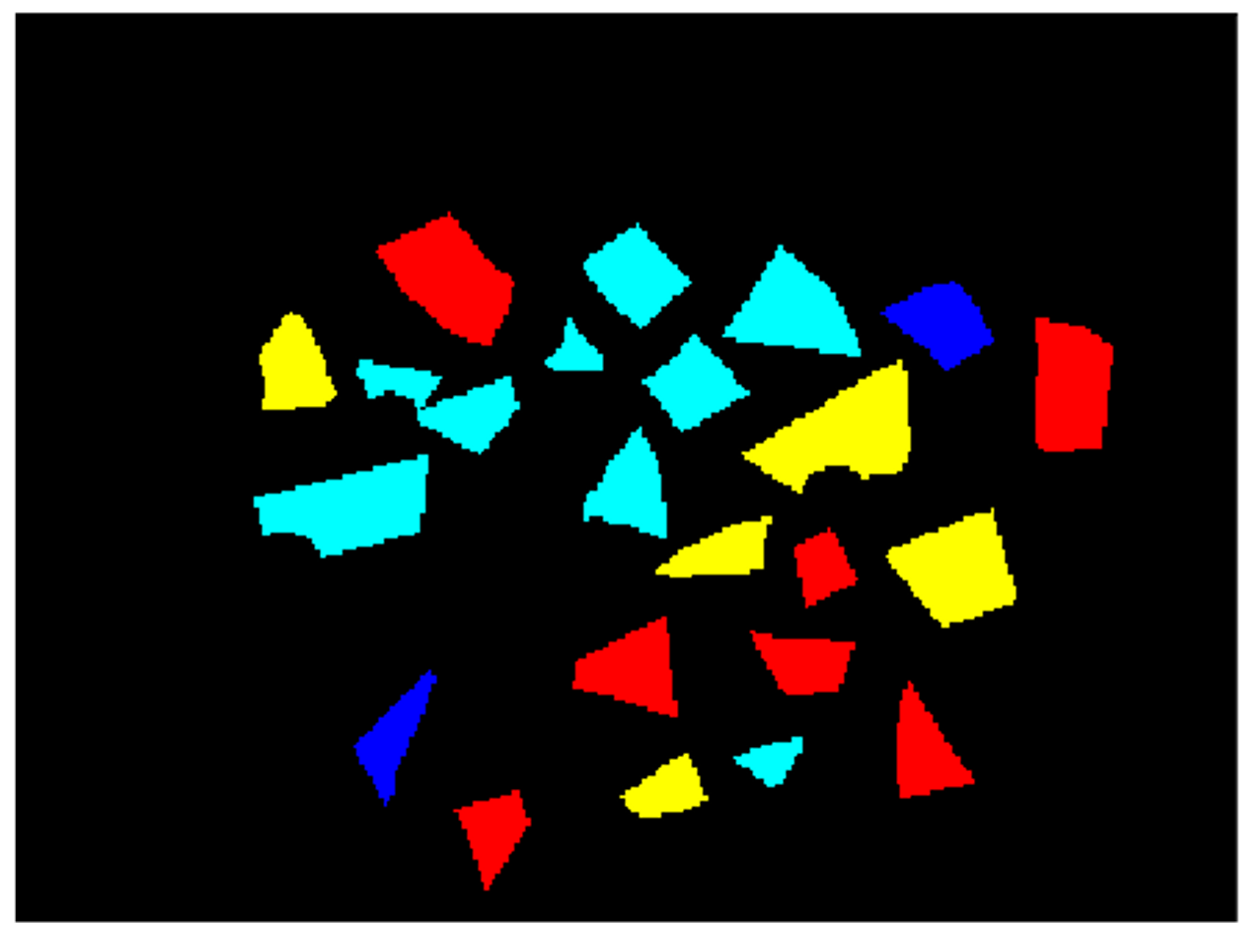}
    \includegraphics[width=\textwidth]{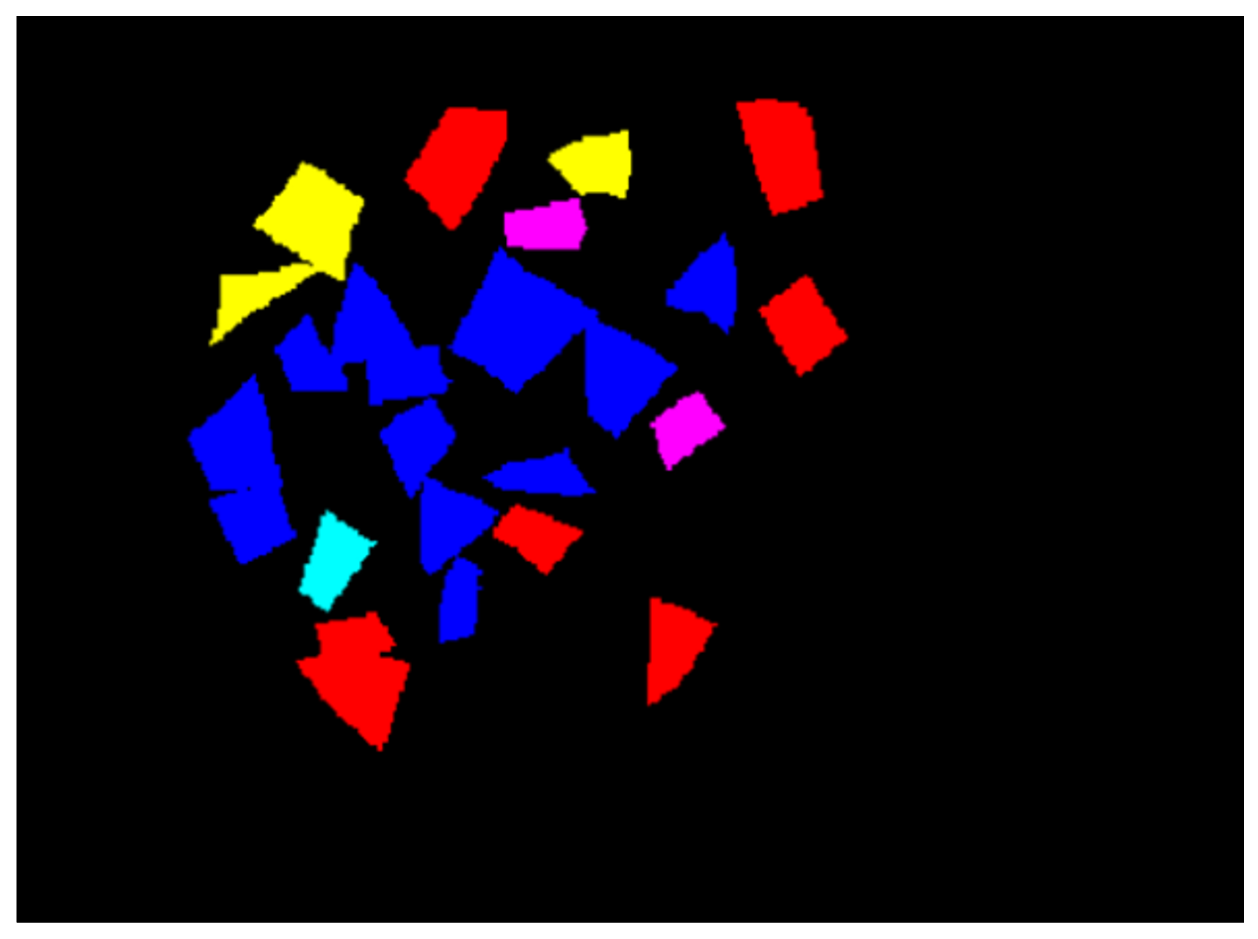}
    \caption{Object classification}
\end{subfigure}
\begin{subfigure}[t]{0.19\textwidth}
    \includegraphics[width=\textwidth]{images/predictions/0/0_gt.png}
    \includegraphics[width=\textwidth]{images/predictions/41/41_gt.png}
    \caption{Ground Truth}
\end{subfigure}
\scriptsize
    \colorbox{Class_1!}{Mesh}
    \colorbox{Class_2!}{Steel-Black}
    \colorbox{Class_3!}{Steel-Gray}
    \colorbox{Class_4!}{HTEL-Anode}
    \colorbox{Class_5!}{HTEL-Cathode}
\caption{Majority voting workflow on sample 1 and sample 42: Instance masks are generated from the RGB images using zero-shot segmentation, then the masks are projected on the pixel-wise classification for approximating the object-wise classification.}
\label{fig:majority_voting_workflow}
\end{figure*}

To better assess material detection performance, we applied an object-wise majority voting strategy based on zero-shot object segmentation. Fig.\ref{fig:majority_voting_workflow} presents the workflow to obtain object-wise classification. Using zero-shot detection, the instance segmentation maps are generated and projected on the pixel-wise classification for approximating the object-wise classification. In this process, objects are represented by polygons, and all pixels within the polygon are assigned the class label that is predicted most frequently among the enclosed pixels. This approach improves classification robustness by incorporating spatial neighborhood information at the object level. Pixel-wise predictions often exhibit noise around object boundaries, mainly due to distorted reflectance signals at the edges where light interacts with slanted or uneven surfaces. This effect is illustrated in Fig. \ref{fig:pixewise predictions}, where central object regions are consistently labeled, while edge regions display higher prediction variability.
The object-wise classification results are presented in Table \ref{tab:majority}. One can observe that majority voting classification consistently improves overall performance across all models when compared to their respective pixel-wise classification results.
Among the evaluated classes, class Mesh was the most accurately detected, followed by Steel Black, Steel Gray, and HTEL Anode. The lowest classification performance was observed for the HTEL Cathode class. These results demonstrate the strong potential of HSI for distinguishing electrolyzers materials and also underscore the need for an extended spectral range to capture more discriminative features, particularly for materials with subtle spectral differences.

\begin{figure}[ht]
    \centering
    \begin{subfigure}{0.7\columnwidth}
        \centering
            \includegraphics[width=\columnwidth]{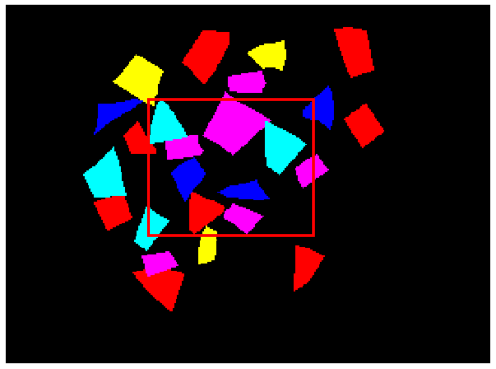}
         \caption{Grund Truth}
           \label{fig:gt}
    \end{subfigure}
    \begin{subfigure}{0.32\columnwidth}
        \centering
        \includegraphics[width=\textwidth]{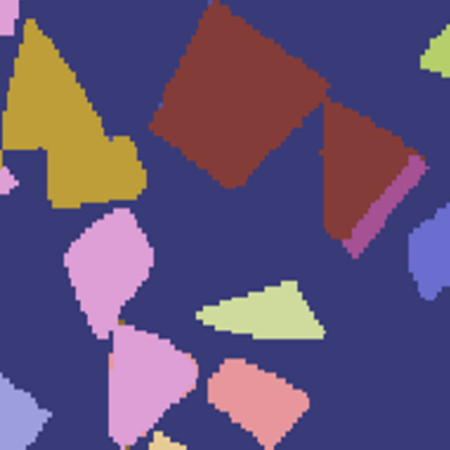}
         \caption{Instance map}
           \label{fig:instance}
    \end{subfigure}
    \begin{subfigure}{0.32\columnwidth}
        \centering
        \includegraphics[width=\textwidth]{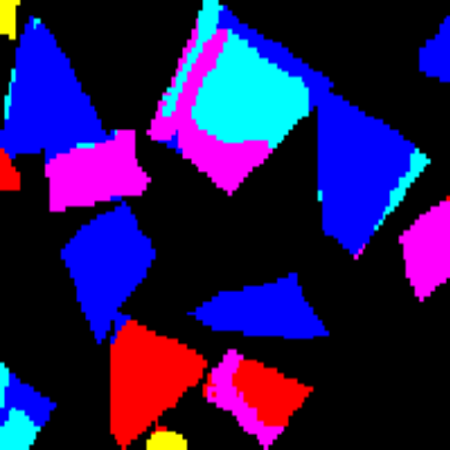}
         \caption{Pixel-wise}
           \label{fig:pixel}
    \end{subfigure}
    \begin{subfigure}{0.32\columnwidth}
        \centering
        \includegraphics[width=\textwidth]{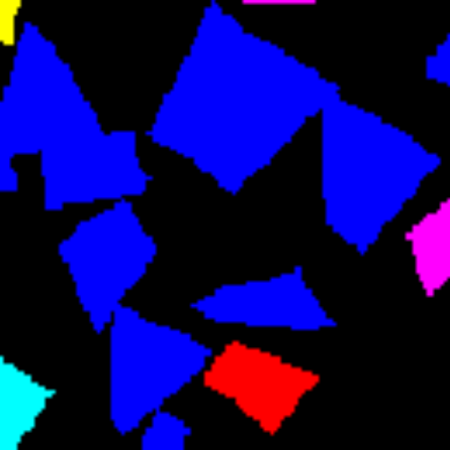}
         \caption{Object-wise}
           \label{fig:voting}
    \end{subfigure}
    \caption{An example case where the  zero-shot object-wise classification fails in sample 42, due to overlapping and touching pieces.}
    \label{fig:object-wise flaws}
\end{figure}

\subsection{Discussion and Future Work}
\begin{figure}[ht]
    \centering
    \begin{subfigure}{0.49\linewidth}
        \centering
            \includegraphics[width=\columnwidth]{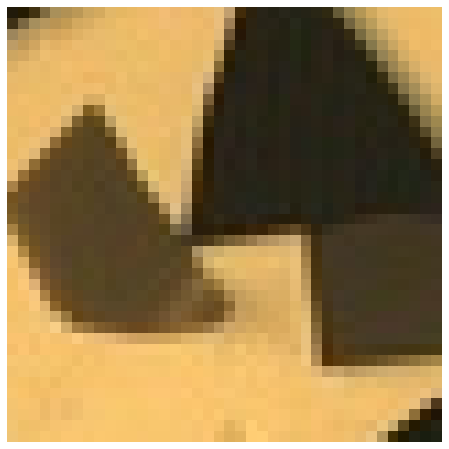} \\
         \caption{Patch image}
           \label{fig:patch_rgb}
    \end{subfigure}
    \begin{subfigure}{0.49\linewidth}
        \centering
        \includegraphics[width=\textwidth]{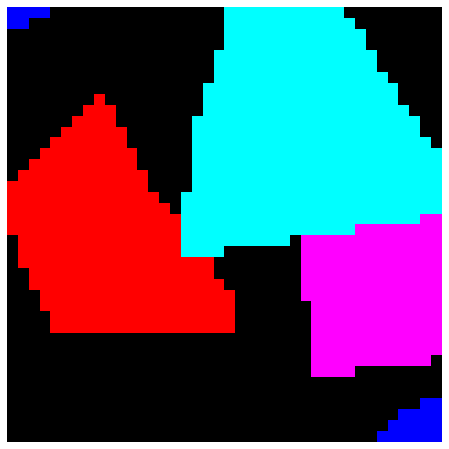}
         \caption{Patch ground truth}
           \label{fig:patch_gt}
    \end{subfigure}
    \caption{Illustration of confusion in HSI Transformer encoders caused by the presence of multiple classes within a single input patch. The patch is taken from test sample 42, Figure. \ref{fig:pixewise predictions}.}
    \label{fig:patch}
\end{figure}
In this subsection, we discuss the challenges along the processing workflow in relation to the HSI Transformer-based models and the zero-shot instance segmentation. We expand further with two major limitations in input data representation and tokenization that affect the performance of Transformer encoders on HSI. First, these models typically classify the center pixel of an HSI patch by treating each spectral band as an individual token. However, because the same spatial patch is used for all wavelengths, the resulting tokens repeat similar spatial patterns with only varying spectral intensities. This redundancy can overload the model with repeated patterns, urging overfitting, and hindering both convergence and generalization.

The second limitation is illustrated in Fig. \ref{fig:patch}, which presents an image patch with mixed-class spectra that negatively affects the models' performance. In the figure,  an input patch is extracted from sample 42 from Fig. \ref{fig:pixewise predictions}. This patch contains a mixture of spectral signatures from three different classes: Mesh, HTEL Anode, and HTEL Cathode. Such mixed-class input introduces complexity, as non-central pixel vectors originating from different materials are merged in the computation layers of the Transformer, potentially disrupting the model’s learning process. The pixel-wise SpectralFormer consistently outperforms its patch-wise counterpart in these scenarios, as it avoids this type of spectral contamination by only processing the spectral vector of the central pixel. This observation underscores the importance of selecting an appropriate patch size based on the spatial scale of the target objects in the HSI to minimize class mixing and improve classification accuracy.

It is clear from Table \ref{tab:majority} that object-wise classification via majority voting on instance segmentation maps improved the performance in general. However, in some cases, it fails. Fig. \ref{fig:object-wise flaws} shows examples of cases where the object-wise classification fails. These cases occur from the direct inference of large zero-shot models pretrained on diverse object datasets \cite{kirillov2023segment}. The SAM model treats  overlapping or touching objects as a single object, leading to incorrect class assignments during majority voting. 
This issue can be addressed by fine-tuning the models on a custom dataset aligned with our object boundary definitions.

Based on the observed performance patterns, we outline several directions for future work aimed at further optimizing Transformer-based models for HSI classification:
\begin{itemize}

    \item \textbf{Object Segmentation Fine-tuning}: The applied zero-shot models should be fine-tuned in order to address the failed cases arising from the direct application of zero-shot models.
    \item \textbf{Refined Input Tokenization}: We aim to explore more advanced tokenization strategies to reduce redundancy and enhance performance.
    \item \textbf{Architectural Enhancement}: We plan to investigate modifications to the model topology, with a specific focus on refining the self-attention mechanism to improve the model's sensitivity to spectral features.
    \item \textbf{Data Engineering}: We plan to conduct additional acquisition scans with diverse samples to effectively reduce model bias, enhance performance in detecting invariant spectral features, and improve the generalizability of SOTA HSI processing models.
    
\end{itemize}
 The data processing and inference pipelines, together with the models' weights, are available on \url{https://github.com/hifexplo}.
\section{Conclusion}
\label{sec:conclusion}

In this study, we introduced Electrolyzers-HSI, a high-resolution multimodal dataset specifically designed to advance the development of smart non-invasive material analysis systems for e-waste recycling. The dataset contains shredded samples of three Electrolyzer materials, captured in the VNIR and SWIR spectral range  (400-2500 nm), with 55 co-registered RGB and HSI images and the classification masks. The dataset's diversity, from single-class to multi-class object configurations, enables comprehensive studies in both controlled and realistic material detection scenarios.
Furthermore, we performed a thorough evaluation of primary Transformer-based models and classical machine learning baselines for pixel-wise and object-level hyperspectral classification. Our findings highlight the advantage of multimodal architectures, with the Multimodal Fusion Transformer consistently outperforming single-modality counterparts by leveraging complementary RGB spatial information. Furthermore, our analysis showed that the pixel-wise SpectralFormer provided more stable performance compared to the patch-based models, as the spectral mixing present in the patch-based input was avoided. This allowed the model to focus on isolated, single-material spectra.
Additionally, to overcome noisy predictions at material boundaries, we integrated zero-shot object segmentation with majority voting, significantly improving the robustness of object-level classification. Plus, we provided the limitations of the used methodologies across multiple steps in the processing pipeline and identified computational bottlenecks in Transformer-based encoders for hyperspectral imaging processing and zero-shot instance segmentation. We proposed future directions to improve the generalization and efficiency of  spectral-spatial feature detection and segmentation performance, supporting technical strategies for industrial-scale implementation. By making Electrolyzers-HSI and our implementations public, we lay a solid foundation for reproducible research and stimulate the development of intelligent, efficient, and scalable material sensing systems to support circular economy initiatives.

{
\small

\bibliographystyle{unsrt}
\bibliography{main}
}

\section{Acknowledgment}
\label{sec:acknowledgment}
The authors express their gratitude to EIT RawMaterials for funding the project 'RAMSES-4-CE' (KIC RM 19262) and BMBF for funding AI4H2 as part of the ReNaRe (3245129018-03HY111D) in the flagship cluster H2Giga". Appreciation is extended to the European Regional Development Fund (EFRE) and the Land of Saxony for their support in funding the computational equipment under the project ’CirculAIre.’ Special thanks go to Yuleika Carolina Madriz Diaz and Filipa Simoes for their assistance in data acquisition.

\section{Author Contributions}
\label{sec:author contributions}
Elias Arbash: conceptualization, methodology, and formal experimental studies and analysis.
Ahmed Jamal Afifi performed the statistical analysis, data preprocessing, and manuscript Documentation.
Ymane Belahsen conducted data acquisition, labelling, and data presentation.
Margret Fuchs, Pedram Ghamisi, Paul Scheunders, and Richard Gloaguen, project management and funds securing, standards and guidance provision.
\section{Competing Interests}
\label{sec:Competing Interests}
The authors declare no competing interests.

\end{document}